\documentclass[journal]{IEEEtran}
\usepackage{amsmath,epsfig}
\usepackage{threeparttable}
\usepackage{multirow}
\usepackage{booktabs}
\usepackage{amssymb}
\usepackage{verbatim}
\usepackage{subfigure}
\usepackage{graphicx}
\usepackage{float}
\usepackage{stfloats}
\usepackage{flushend}
\usepackage{cite}

\renewcommand{\thesubfigure}{\roman{subfigure}} \makeatletter 
\renewcommand{\@thesubfigure}{(\thesubfigure)\space} 
\renewcommand{\p@subfigure}{\thefigure} \makeatother

\begin{document}

\title{A Novel Perspective to Zero-shot Learning: \\Towards an Alignment of Manifold Structures \\
via Semantic Feature Expansion}

\author{Jingcai~Guo,~\IEEEmembership{Member,~IEEE,}
        and~Song~Guo,~\IEEEmembership{Fellow,~IEEE}

%\thanks{Manuscript received August 3, 2019; revised December 30, 2019 and March 1, 2020; accepted March 20, 2020.}

%\thanks{This research was financially supported by the National Natural Science Foundation of China (Grant 61872310) and the Innovation and Technology Commission of the HKSAR to the Hong Kong Branch of National Rail Transit Electrification and Automation Engineering Technology Research Center (No. BBV2). The associate editor coordinating the review of this manuscript and approving it for publication was Dr. Zhu Liu. \textit{(Corresponding author: Song Guo.)}}

%\thanks{J. Guo and S. Guo are with the Department
%of Computing, The Hong Kong Polytechnic University, Hong Kong SAR, China (e-mail: cscjguo@comp.polyu.edu.hk, song.guo@polyu.edu.hk).}
}

%\markboth{IEEE TRANSACTIONS ON MULTIMEDIA,~Vol.~**, No.~**, MARCH~2020}%

\maketitle

\begin{abstract}
Zero-shot learning aims at recognizing unseen classes (no training example) with knowledge transferred from seen classes. This is typically achieved by exploiting a semantic feature space shared by both seen and unseen classes, i.e., attribute or word vector, as the bridge. One common practice in zero-shot learning is to train a projection between the visual and semantic feature spaces with labeled seen classes examples. When inferring, this learned projection is applied to unseen classes and recognizes the class labels by some metrics. However, the visual and semantic feature spaces are mutually independent and have quite different manifold structures. Under such a paradigm, most existing methods easily suffer from the domain shift problem and weaken the performance of zero-shot recognition. To address this issue, we propose a novel model called AMS-SFE. It considers the alignment of manifold structures by semantic feature expansion. Specifically, we build upon an autoencoder-based model to expand the semantic features from the visual inputs. Additionally, the expansion is jointly guided by an embedded manifold extracted from the visual feature space of the data. Our model is the first attempt to align both feature spaces by expanding semantic features and derives two benefits: first, we expand some auxiliary features that enhance the semantic feature space; second and more importantly, we implicitly align the manifold structures between the visual and semantic feature spaces; thus, the projection can be better trained and mitigate the domain shift problem. Extensive experiments show significant performance improvement, which verifies the effectiveness of our model.
\end{abstract}


%\IEEEpeerreviewmaketitle

\section{Introduction}

\begin{figure}[t]
    \centerline{\includegraphics[width=0.48\textwidth]{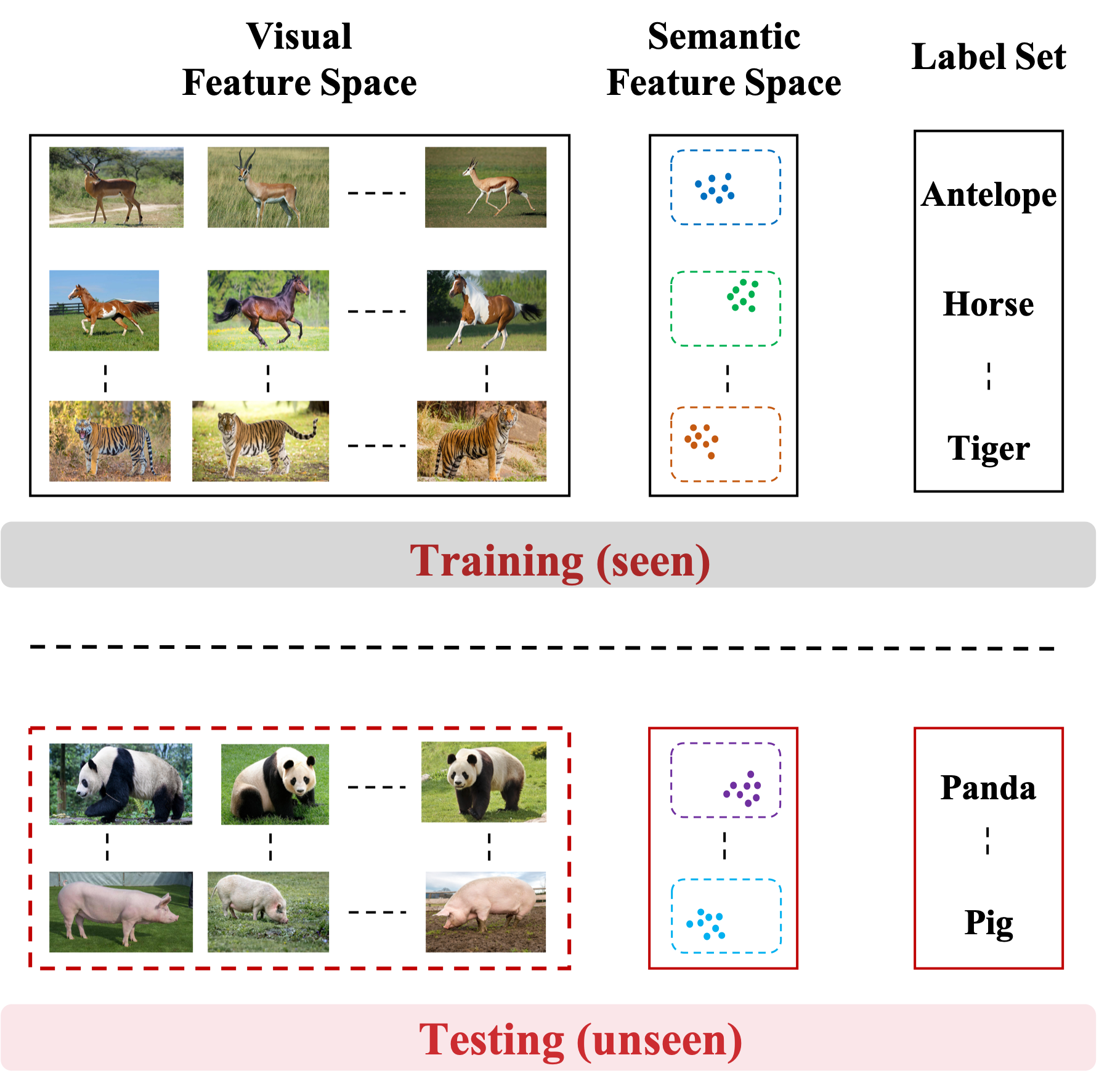}}
    \caption{An illustration of zero-shot learning: the training is conducted solely on the seen class and the testing is conducted on unseen classes. The unseen class examples, i.e., bounded with a red dotted box, are not available during the whole training process. The two label sets are disjoint from each other and are only bridged by the semantic feature space.}
    \label{fig1}
\end{figure}

\IEEEPARstart{I}{n} the past few years, with the increasing development of deep learning techniques, many machine learning tasks and techniques have been proposed and consistently achieved state-of-the-art performance. Among them, most of the tasks can be grouped into supervised learning problems, such as image classification \cite{krizhevsky2012imagenet,zhao2017diversified,zhang2018multilabel}, face verification \cite{hu2014discriminative,sun2014deep,ding2015robust}, multimedia retrieval \cite{rasiwasia2010new,kang2015learning}, medical imaging \cite{greenspan2016guest}, and financial forecasting \cite{ding2015deep}. These tasks usually require a large number of labeled examples to train the model and then to make inferences on testing examples. In the implementation of deep learning techniques, the risk of overfitting problems that commonly exist in classical machine learning models, e.g., support vector machines \cite{scholkopf2001learning,guo2016improved} and tree/forest-based models \cite{liaw2002classification}, can be offset by the depth and width of deep neural networks. Generally, in most cases, the larger the quantity of data, the better the model performance. Taking ImageNet \cite{deng2009imagenet} as an example, which consists of 21,841 classes and 14,197,122 images in total, its emergence has brought unprecedented opportunities to the computer vision area. Many tasks trained on large datasets, e.g., ImageNet, and related tasks with transfer learning, e.g., pretraining models on ImageNet, continue to make progress in contrast to many other areas. These tasks have achieved superior performance and even surpassed humans in some scenarios \cite{he2015delving}. Despite the great power of supervised learning, it relies too much on a large number of labeled data examples, which makes it difficult for models to be generalized to unfamiliar or even unseen classes. Transfer learning \cite{torrey2010transfer} has partially solved this problem; it pretrains a model on a source domain dataset of a similar task and then transfers the whole or part of the trained model to the target domain task and fine-tunes with target data. For example, it is easier for a learner who has already learned English to learn French because many internal similarities and overlaps exist between these two languages.

Humans have an excellent ability to generalize learned knowledge to explore the unknown. It has been proven that humans can recognize over 30,000 object classes and many more subclasses \cite{biederman1987recognition}, e.g., breeds of birds and combinations of attributes and objects. Moreover, humans are very good at recognizing object classes without previously seeing their real examples. For example, if a learner who has never seen a panda is taught that the panda is a bear-like animal that has black and white fur, then he or she will be able to easily recognize a panda when seeing a real example. In machine learning, this is considered the problem of \textit{zero-shot learning} (ZSL) \cite{lampert2009learning}. The setting of ZSL can be regarded as an extreme case of transfer learning: the model is trained to imitate human ability in recognizing examples of unseen classes that are not shown during the training stage \cite{frome2013devise,lampert2014attribute,shigeto2015ridge,wang2015zero,bucher2016improving,zhang2016zero,changpinyo2016synthesized,kodirov2017semantic,zhu2018generative,guo2019ams}. ZSL is typically achieved by taking the utilization of labeled seen class examples and certain \textit{common knowledge} that can be shared and transferred between seen and unseen classes. This common knowledge is per-class, semantic and high-level features for both the seen and unseen classes \cite{lampert2009learning}, which enables easy and fast implementation and inference to ZSL. Among them, semantic attributes and semantic word vectors have become the most popular common knowledge in recent years. The semantic attributes are meaningful high-level information about examples, such as shape, color, component, and texture. In contrast, semantic word vectors are vector representations of words learned from a large external corpus, of which two word vectors are expected to have a small distance if two words more frequently appear in context than others. As such, similar classes should have similar patterns in the semantic feature space, and this particular pattern is defined as the prototype. Each class is embedded in the semantic feature space and endowed with a prototype, so the ZSL can be easily extended to much broader classes by collecting class-level prototypes instead of example-level collecting and labeling, which are expensive and time consuming.

In conventional supervised learning (CSL), the training and testing examples belong to the same class-set, which means that the learned model has already seen some examples of all the classes it encounters during testing. In contrast, the ZSL only trains the model on seen class examples, and the learned model is expected to infer novel/unseen class examples. Thus, the essential difference between the ZSL and CSL is that the training and testing class sets of ZSL are disjoint from each other. As a result, the ZSL can be regarded as a complement to the CSL and can handle some scenarios where labeled data are scarce or difficult to obtain for some classes. Taking image classification/recognition as an example, even some of the largest datasets, e.g., ImageNet, cannot cover all the classes in real-world applications, and some classes that are only composed of very few or even only one real example may also exist. Collecting and labeling the examples of all classes is impossible. However, recognizing as many classes as possible is urgently needed in real-world applications. As such, the ZSL, which is capable of transferring knowledge from seen to unseen classes has received increasing attention in recent years \cite{frome2013devise,lampert2014attribute,shigeto2015ridge,wang2015zero,bucher2016improving,zhang2016zero,changpinyo2016synthesized,kodirov2017semantic,zhu2018generative,guo2019ams}. An illustration of the basic settings of zero-shot learning is shown in Fig. \ref{fig1}

As a common practice in ZSL, an unseen class example is first projected from the original input feature space, i.e., the visual feature space, to the semantic feature space by a projection trained on seen classes. Then, with such obtained semantic features, we search the most closely related prototype whose corresponding class is set to this example. Specifically, this relatedness can be measured by metrics such as the similarity or distance between the semantic features and prototypes. Thus, some simple algorithms, such as nearest-neighbor (NN), can be applied to search the class prototypes. However, due to the absence of unseen classes when training the projection, the \textit{domain shift} problem \cite{fu2015transductive} easily occurs. This is mainly because the visual and semantic feature spaces are mutually independent. More specifically, visual features represented by high-dimensional vectors are usually not semantically meaningful, and the semantic features are also often not visually meaningful. Therefore, it is challenging to obtain a well-matched projection between the visual and semantic feature spaces.

\begin{figure*}[t]
    \centerline{\includegraphics[width=0.885\textwidth]{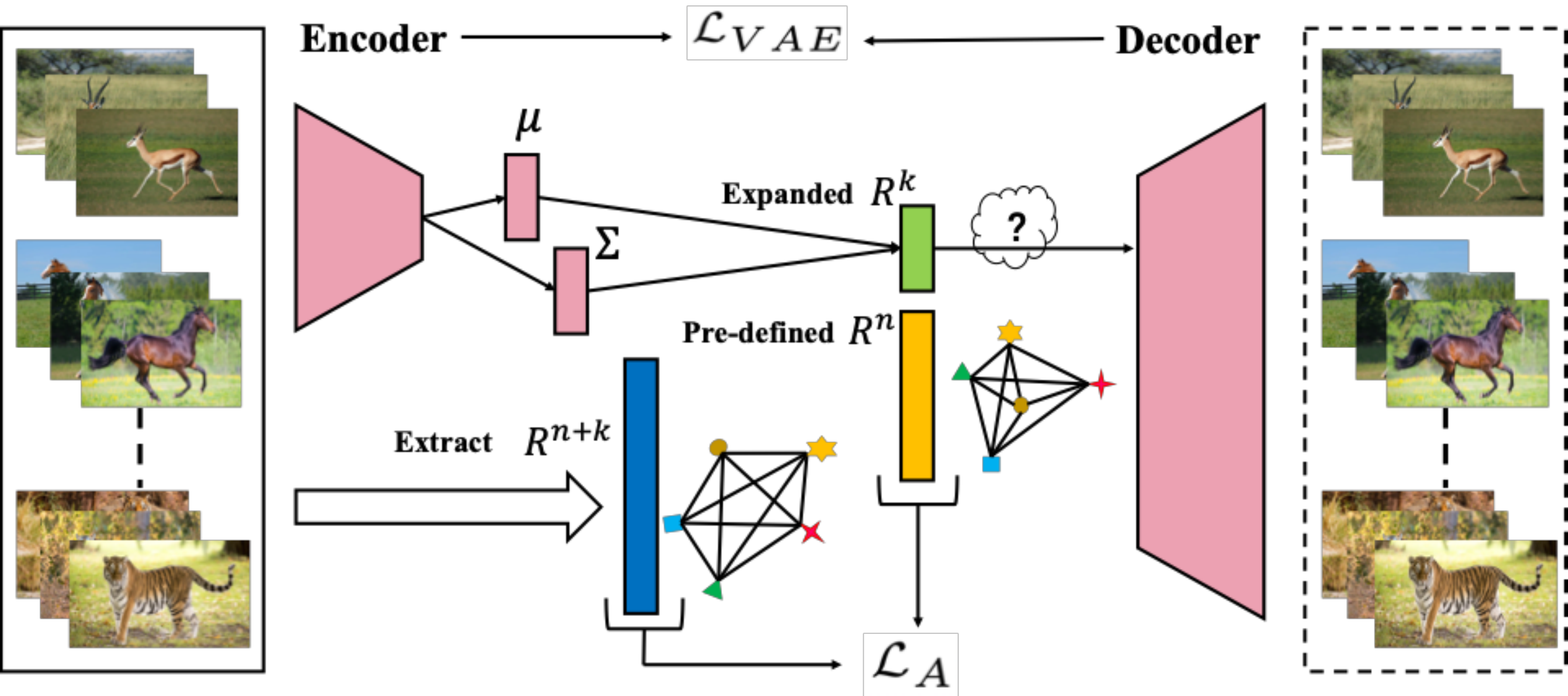}}
    \caption{The proposed AMS-SFE expands the semantic feature space to implicitly align the manifold structures between the visual and semantic feature spaces. Our model can be roughly divided into 3 steps: (1) expand several auxiliary semantic features (green) except for the predefined features (yellow); (2) extract a low-dimensional embedded manifold of the original visual feature space, which retains the geometrical and structural information of the visual features of data (blue); and (3) combine these two semantic features and force the manifold structure to approximate the structure of the embedded manifold extracted from the original visual feature space. The expansion and alignment are jointly achieved within the training of the VAE.}
    \label{fig2}
\end{figure*}

To address the above issues, we propose a novel model to align the manifold structures between the visual and semantic feature spaces, as shown in Fig. \ref{fig2}. Specifically, we train an autoencoder-based model that takes the visual features as input to generate $k$-dimensional auxiliary features for each prototype in the semantic feature space except for the predefined $n$-dimensional features. Additionally, we combine these auxiliary semantic features with the predefined features to discover better adaptability for the semantic feature space. This adaptability is mainly achieved by aligning the manifold structures between the combined semantic feature space ($\mathbb{R}^{n+k}$) to an embedded $(n+k)$-dimensional manifold extracted from the original visual feature space of data. The expansion and alignment phases are conducted simultaneously by joint supervision from both the reconstruction and alignment terms within the autoencoder-based model. Our model results in two benefits: (1) we can enhance the representation capability of semantic feature space, so it can better adapt to unseen classes; (2) we can implicitly align the manifold structures between the visual and semantic feature spaces, so the domain shift problem can be better mitigated. Our contributions are three-fold:
\begin{itemize}
\item We are the first to consider the expansion of semantic feature space for zero-shot learning and align between the visual and semantic feature spaces.
\item Our model obtains a well-matched visual-semantic projection that can mitigate the domain shift problem.
\item Our model outperforms various existing representative methods with significant improvements and shows its effectiveness.

\end{itemize}

The rest of this paper is organized as follows. Section II introduces the related work. Then, in Section III, we present our proposed method. Section IV discusses the experiment, and the conclusion is addressed in Section V.

\section{Related Work}
In this section, we mainly introduce some related problems and methods concerned with this paper.

\subsection{Visual-Semantic Projection}
Existing ZSL methods have established three main directions for visual-semantic projection: forward, reverse and intermediate projections.

\subsubsection{Forward Projection}
The forward projection is the most widely used projection in ZSL. It refers to finding a projection that maps the visual feature space to the semantic feature space. SOC \cite{palatucci2009zero} maps the visual features to the semantic feature space and then searches the nearest class embedding vector. SJE \cite{akata2015evaluation} optimizes the structural SVM loss to learn the bilinear compatibility, while ESZSL \cite{romera2015embarrassingly} utilizes the square loss to learn the bilinear compatibility and adds a regularization term to the objective with respect to the Frobenius norm. ALE \cite{akata2016label} trains a bilinear compatibility function between the semantic attribute space and the visual space by ranking loss. Similarly, \cite{frome2013devise} also trained a linear mapping function between visual and semantic feature space by an efficient ranking loss formulation. \textit{Bucher et al.} \cite{bucher2016improving} embedded the visual features into the attribute space. Recently, SAE \cite{kodirov2017semantic} used a semantic autoencoder to regularize zero-shot recognition. \textit{Xian et al.} \cite{xian2016latent} extended the bilinear compatibility model of SJE \cite{akata2015evaluation} to be a piecewise linear model by learning multiple linear mappings with the selection being a latent variable. \textit{Socher et al.} \cite{socher2013zero} used a deep learning model that contains two hidden layers to learn a nonlinear mapping from the visual feature space to the semantic word vector space \cite{mikolov2013distributed}.

\subsubsection{Reverse Projection}
In contrast, with forward projection, some ZSL models aim to find a projection that reversely maps the semantic feature space back to the visual feature space. \textit{Ba et al.} \cite{ba2015predicting} and \textit{Zhang et al.} \cite{zhang2017learning} both train a deep neural network to map the semantic features to the visual feature space. \textit{Changpinyo et al.} \cite{changpinyo2016predicting} proposed a simple model based on a support vector regressor to map the semantic features to the visual feature space and performed nearest-neighbor algorithms.

\subsubsection{Intermediate Projection}
The intermediate projection refers to finding an intermediary feature space that both the visual features and the semantic features are mapped to \cite{lu2015unsupervised}. \textit{Zhang et al.} \cite{zhang2015zero} utilized the mixture of seen class parts as the intermediate feature space; then, the examples belonging to the same class should have similar mixture patterns. \textit{Zhang et al.} \cite{zhang2016zero} maps the visual features and semantic features to two separate intermediate spaces. Additionally, some researchers also propose several hybrid models to jointly embed several kinds of textual features and visual features to ground attributes \cite{fu2015zero,akata2016multi,long2017zero,changpinyo2016synthesized}. \textit{Lu et al.} \cite{lu2015unsupervised} linearly combines the base classifiers trained in a discriminative learning framework to construct the classifier of unseen classes.

In our model, we mainly consider the performance on the forward projection. Visual examples are mapped to the semantic feature space by the learned projection and then search for the corresponding prototypes to determine their classes.

\subsection{Domain Shift Problem}
The domain shift problem was first identified and studied by \textit{Fu et al.} \cite{fu2015transductive}. It refers to the phenomenon that when projecting unseen class examples from the visual feature space to the semantic feature space, the obtained results may shift away from the real results (prototypes). The domain shift problem is essentially caused by the nature of ZSL that the training (seen) and testing (unseen) classes are mutually disjoint. Recently, several researchers have investigated how to mitigate the domain shift problem, including \textit{inductive learning}-based methods, which enforce additional constraints from the training data \cite{fu2015zero,changpinyo2016synthesized}, and \textit{transductive learning}-based methods, which assume that the unseen class examples (unlabeled) are also available during training \cite{fu2015transductive,li2017zero,song2018transductive}. 

It should be noted that the model performance of the transductive setting is generally better than that of the inductive setting because of the utilization of extra information from unseen classes during training, thus naturally avoiding the domain shift problem. However, transductive learning does not fully comply with the \textit{zero-shot} setting in which no examples from an unseen class are available. With the popularity of generative adversarial networks (GANs), some related ZSL methods have also been proposed recently. GANZrl \cite{AAAI1816805} applied GANs to synthesize examples with specified semantics to cover a higher diversity of seen classes. In contrast, GAZSL \cite{zhu2018generative} leverages GANs to imagine unseen classes from text descriptions. Although several works have already achieved some progress, the domain shift problem is still an open issue. In our model, the expansion phase is also a generative task but focuses on the semantic feature level. We adopt an autoencoder-based model that is lighter and easier to implement yet effective. Moreover, we strictly comply with the \textit{zero-shot} setting and isolate all unseen class examples from the training process.

\subsection{Manifold Learning}
The manifold is a concept from mathematics that refers to a topological space that locally resembles Euclidean space near each point. Manifold learning is based on the idea that there exists a lower-dimensional manifold embedded in a high-dimensional space \cite{wang2017quantifying}. Recently, some manifold learning-based ZSL models have been proposed. \textit{Fu et al.} introduces the semantic manifold distance to redefine the distance metric in the semantic feature space using an absorbing Markov chain process. MFMR \cite{xu2017matrix} leverages the sophisticated technique of matrix trifactorization with manifold regularizers to enhance the projection between the visual and semantic spaces. In our model, we consider obtaining an embedded manifold in a lower-dimensional space of data in the original visual feature space. This embedded manifold is expected to retain the geometrical and distribution constraints in the visual feature space. Such manifold information is further used to guide the alignment of manifold structures between the visual and semantic feature spaces.

\section{Proposed Method}
In this section, we first give the formal problem definition of zero-shot learning. Next, we introduce our proposed method and formulation in detail. More specifically, an autoencoder-based network is first applied to generate and expand some auxiliary semantic features except for the predefined ones. Then, we extract a lower-dimensional embedded manifold from the original visual feature space of the data, which properly retains its geometrical and distribution constraints. By using the obtained embedded manifold, we then construct an additional regularization term to guide the alignment of manifold structures between the visual and semantic feature spaces. Finally, the prototype updating strategy and the recognition process of our model are addressed.

\subsection{Problem Definition}
We start by formalizing the zero-shot learning task and then introduce our proposed method and formulation. Given a set of labeled seen class examples $\mathcal{D}=\left \{ x_{i}, y_{i} \right \}_{i=1}^{l}$, where $x_{i}\in\mathbb{R}^{d}$ is a seen class example, i.e., visual features, with class label $y_{i}$ belonging to $m$ seen classes $C=\left \{ c_{1}, c_{2}, \cdots , c_{m}\right \}$, the goal is to construct a model for a set of unseen classes ${C}' = \left \{ {c}'_{1}, {c}'_{2}, \cdots , {c}'_{v}\right \}$ ($C \bigcap {C}' = \phi$) that have no labeled examples during training. In the testing phase, given a test example ${x}'\in\mathbb{R}^{d}$, the model is expected to predict its class label $c({x}')\in{C}'$. To this end, some side information, i.e., the semantic features, denoted as $S^{p} = \left ( a_{1}, a_{2}, \cdots , a_{n} \right )\in \mathbb{R}^{n}$, is needed as common knowledge to bridge the seen and unseen classes in the semantic feature space, where each $a_{i}$ is one feature dimension in such semantic feature space. Therefore, the seen class examples $\mathcal{D}$ can be further specified as $\mathcal{D}=\left \{ x_{i}, y_{i}, S^{p}_{i} \right \}_{i=1}^{l}$. Each seen class $c_{i}$ is endowed with a predefined semantic prototype $P^{p}_{c_{i}}\in \mathbb{R}^{n}$, and each unseen class ${c_{i}}'$ is also endowed with a predefined semantic prototype ${P^{p}_{{c_{i}}'}}'\in \mathbb{R}^{n}$. Thus, for each seen class example, we have $S^{p}_{i} \in P^{p}=\left \{P^{p}_{c_{1}}, P^{p}_{c_{2}}, \cdots , P^{p}_{c_{m}}  \right \}$, while for unseen class example ${x}'$, we need to predict its semantic features ${S^{p}}' \in \mathbb{R}^{n}$ and set the class label by searching the most closely related semantic prototype within ${P^{p}}'=\left \{{P^{p}_{{c_{1}}'}}', {P^{p}_{{c_{2}}'}}', \cdots , {P^{p}_{{c_{v}}'}}'  \right \}$.

\subsection{Method and Formulation}

\subsubsection{Semantic Feature Expansion}
To align the manifold structures between the visual and semantic feature spaces, the first step from the bottom up is to expand the semantic features. Specifically, we keep the predefined $n$-dimensional semantic features $S^{p}=(a_{1}, a_{2}, \cdots , a_{n})\in \mathbb{R}^{n}$ fixed and expand extra $k$-dimensional auxiliary semantic features $S^e = \left (a_{n+1}, a_{n+2}, \cdots , a_{n+k}\right )\in \mathbb{R}^{k}$ to enlarge the target semantic feature space as $\mathbb{R}^{n+k}$. Practically, several techniques, e.g., the generative adversarial network (GAN) \cite{goodfellow2014generative} and the autoencoder (AE) \cite{baldi2012autoencoders}, can achieve this target on the basis that the expanded auxiliary semantic features are faithful to the original input features, i.e., visual features. Compared with the GAN, which is good at synthesizing data distribution, e.g., more realistic images in example-level \cite{zhang2018stackgan++}, the AE is much lighter and easier to train \cite{vincent2010stacked,yu2013embedding,chu2017stacked,guo2019ee}. Moreover, in our model, the expanded auxiliary features are expected to be more per-class semantically high-level in contrast to example-level features. Considering the generation target, the training cost and the model complexity, the AE is a better choice for our purpose. The standard AE consists of two parts: the encoder maps the visual features $x \in \mathbb{R}^{d}$ to a latent feature space, in which the latent features $z \in \mathbb{R}^{k} (k\ll d)$ are normally a high-level and compact representation of the visual features. The decoder then maps the latent features back to the visual feature space and reconstructs the original visual features as $\hat{x} \in \mathbb{R}^{d}$. The AE loss measuring the reconstruction can be described as:
\begin{equation}
\mathcal{L}_{AE} = \sum_{x\sim \mathcal{D}}\left \| x - \hat{x} \right \|_{2}^{2}\,.
\end{equation}
We minimize the objective of Eq. (1) to guarantee the learned latent features $z_{i}$ retain the most powerful information of the input $x_{i}$. It should be noted that the SAE \cite{kodirov2017semantic} first implemented the AE for the zero-shot learning task, which applies the semantic autoencoder training framework to obtain the projection between the visual and semantic feature spaces. However, in our model, the AE is mainly implemented in the expansion process to obtain extra auxiliary semantic features.

The standard AE performs well in our model despite the latent feature space being pointwise sensitive, which means that the margins of each class within the latent feature space are discrete and the data in the latent feature space are unevenly distributed. In other words, some areas in the latent feature space do not represent any data. Although the prototypes of each class are also inherently discrete, a smooth and continuous latent feature space can intuitively better represent the margins among classes and makes the projection between the visual and latent feature spaces more robust. To this end, we further apply the variational autoencoder (VAE) \cite{DBLP:journals/corr/KingmaW13} to formulate the expansion process in our model.

Different from the standard AE, the encoder of the VAE predicts the mean feature vector $\mu$ and the variance matrix $\Sigma$, such that the distribution of latent features $q \left ( z |x \right )$ can be approximated by $\mathcal{N} \left ( \mu, \Sigma \right )$, i.e., $q \left ( z |x \right ) = \mathcal{N} \left ( \mu, \Sigma \right )$, from which a latent feature $z$ is sampled and further decoded to reconstruct the original visual features as $\hat{x} \in \mathbb{R}^{d}$. The key difference between the AE and VAE is the embedding methods of the inputs in the latent feature space. The AE learns a compressed data representation that is normally more discrete, while the VAE attempts to learn the parameters of a probability distribution representing the data, which makes the learned latent features smoother and more continuous \cite{DBLP:journals/corr/KingmaW13,blei2017variational}. However, the sampling operation from $\mathcal{N} \left ( \mu, \Sigma \right )$ is nondifferentiable and makes backpropagation impossible. As suggested by the reparameterization trick \cite{DBLP:journals/corr/KingmaW13}, sampling from $z \sim \mathcal{N} \left ( \mu, \Sigma \right )$ is equivalent to sampling $\epsilon \sim \mathcal{N} \left ( 0, I \right )$ and setting $z = \mu + \Sigma ^{\frac{1}{2}} \epsilon$. Thus, $\epsilon$ can be regarded as an input of the network and makes the sampling operation differentiable. Moreover, we need an additional constraint other than the reconstruction loss, i.e., $\mathcal{L}_{AE}$, to guide the training of VAE. It should be noted that the additional constraint is expected to force the latent feature distribution to be similar to a prior, so the objective of the VAE can be further specified as:
\begin{equation}
\mathcal{L}_{VAE} = \sum_{x\sim \mathcal{D}}\left \| x - \hat{x} \right \|_{2}^{2}-D_{KL}  \big (q\left ( z | x \right )  \Vert p\left ( z \right ) \big ),
\end{equation}
where the first term is the conventional reconstruction loss, which forces the latent feature space to be faithful and restorable to the original visual feature space, and the second term is the unpacked Kullback-Leibler divergence between the latent feature and the chosen prior $p\left ( z \right )$, e.g., a multivariate standard Gaussian distribution, which further forces the margins of each class to be smooth and continuous and makes the visual-semantic projection more robust.

\subsubsection{Embedded Manifold Extraction}
Before exploiting these auxiliary semantic features, we need to extract a lower-dimensional embedded manifold ($\mathbb{R}^{n+k}$) of the visual feature space ($\mathbb{R}^{d},n+k\ll d$) to utilize the structure information. To this end, we first find and define the center of each seen class in the visual feature space as $x^{c} = \left \{x^{c_{i}} \right \}_{i=1}^{m}$, where $\left \{c_{i} \right \}_{i=1}^{m}$ are $m$ class labels and $x^{c_{i}}$ is the center, e.g., the mean value, of all examples belonging to class $c_{i}$. We compose a matrix $\mathbf{D} \in \mathbb{R}^{m\times m}$ to record the distance of each center pair from $x^{c}$ in the original visual feature space as:
\begin{equation}
\mathbf{D} = \begin{bmatrix}
d_{1,1} & d_{1,2} & \cdots  & d_{1,m}\\ 
d_{2,1} & d_{2,2} & \cdots  & d_{2,m} \\  
\vdots  & \vdots  &   & \vdots \\ 
d_{m,1} & d_{m,2} & \cdots  & d_{m,m} 
\end{bmatrix},
\end{equation}
where $d_{i,j}$ is calculated as $\left \| x^{c_{i}} - x^{c_{j}} \right \|$. Then, our target is to search for a lower-dimensional embedded manifold ($\mathbb{R}^{n+k}$) that can be modeled by an $(n+k)$-dimensional embedded feature representation, i.e., denoted as $\mathbf{O}=\left [ o_{i} \right ]\in \mathbb{R}^{(n+k) \times m}$, where each $o_{i}$ is the embedded representation of the class center $x^{c_{i}}$. The embedded representation is expected to retain the geometrical and structural information of the visual feature space of the data.

To obtain $\mathbf{O}$, a natural and straightforward approach is that the distance matrix $\mathbf{D}$ can also restrain the embedded representation $\mathbf{O}$, which means that the distance of each point pair of $\mathbf{O}$ also has the same distance matrix $\mathbf{D}$ in the corresponding $(n+k)$-dimensional feature space. To solve this problem, we denote the inner product of $\mathbf{O}$ as $\mathbf{B} = \mathbf{O}^\top \mathbf{O}\in \mathbb{R}^{m \times m}$, so that $b_{ij} = o_i^\top o_j$ and we obtain:
\begin{equation}
\begin{aligned}
d_{ij}^2 &= \left \| o_i \right \|^2 + \left \| o_j \right \|^2-2o_i^\top o_j = b_{ii} + b_{jj} - 2b_{ij}.
\end{aligned}
\end{equation}
We set $\sum_{i=1}^{m}o_{i}=0$ to simplify the problem so that the summation of each row/column of $\mathbf{O}$ equals zero. The zero-centered setting can reduce computations while retaining the data's geometrical and structural information. Then, we can easily obtain:
\begin{equation}
\sum_{i=1}^{m}d_{ij}^{2} = \mathrm{Tr}(\mathbf{B}) + mb_{jj}, 
\end{equation}
\begin{equation}
\sum_{j=1}^{m}d_{ij}^{2} = \mathrm{Tr}(\mathbf{B}) + mb_{ii}, 
\end{equation}
\begin{equation}
\sum_{i=1}^{m}\sum_{j=1}^{m}d_{ij}^{2} = 2m\mathrm{Tr}(\mathbf{B}), 
\end{equation}
where $\mathrm{Tr}(\cdot)$ is the trace of the matrix, i.e., $\mathrm{Tr}(B) = \sum_{i=1}^{m}\left \| o_i \right \|^2$. We denote:
\begin{equation}
d_{i\cdot }^{2} = \frac{1}{m}\sum_{j=1}^{m}d_{ij}^2,
\end{equation}
\begin{equation}
d_{\cdot j}^{2} = \frac{1}{m}\sum_{i=1}^{m}d_{ij}^2,
\end{equation}
\begin{equation}
d_{\cdot \cdot}^{2} = \frac{1}{m^2}\sum_{i=1}^{m}\sum_{j=1}^{m}d_{ij}^2.
\end{equation}
From Eq. (4), we can easily obtain:
\begin{equation}
b_{ij} = -\frac{1}{2}\left ( d_{ij}^{2} - b_{ii} - b_{jj} \right ).
\end{equation}
From Eqs. (7) and (10), we can obtain:
\begin{equation}
\mathrm{Tr}(\mathbf{B}) = \frac{1}{2m}\sum_{i=1}^{m}\sum_{j=1}^{m}d_{ij}^2 = \frac{1}{2}md_{\cdot \cdot}^{2},
\end{equation}
From Eqs. (6) and (8), and Eqs. (5) and (9), respectively, we can obtain:
\begin{equation}
\left\{
\begin{array}{lr}
b_{ii} = \frac{1}{m}\sum_{j=1}^{m}d_{ij}^2 - \frac{1}{m}\mathrm{Tr}(\mathbf{B})=d_{i \cdot}^{2} - \frac{1}{2}d_{\cdot \cdot}^{2}  &  \\
b_{jj} = \frac{1}{m}\sum_{i=1}^{m}d_{ij}^2 - \frac{1}{m}\mathrm{Tr}(\mathbf{B})=d_{\cdot j}^{2} - \frac{1}{2}d_{\cdot \cdot}^{2} &  
\end{array},
\right.
\end{equation}
Combining Eqs. (11)$\sim$(13), we can obtain the inner product matrix $\mathbf{B}$ by the distance matrix $\mathbf{D}$ as:
\begin{equation}
b_{ij} = -\frac{1}{2}\left ( d_{ij}^2 - d_{i\cdot }^2 - d_{\cdot j}^2 + d_{\cdot \cdot }^2 \right ).
\end{equation}
By applying eigenvalue decomposition (EVD) \cite{chonavel2003fast} with $\mathbf{B}$, we can easily obtain the $(n+k)$-dimensional representation $\mathbf{O}$, which models the geometrical and structural information of the expected $(n+k)$-dimensional embedded manifold ($\mathbb{R}^{n+k}$).

\subsubsection{Manifold Structure Alignment}
With the obtained $\mathbf{O}$, we can now align the manifold structures between the visual and semantic feature space. Specifically, we measure the similarity of the combined semantic feature representation $S^{p+e}$ (predefined $S^p$ combined with expanded $S^e$) and the embedded representation $\mathbf{O}$ by cosine distance, in which the output similarity between two vectors is bounded from -1 to 1 and is magnitude free. It should be noted that in our model, the alignment is jointly completed with the semantic feature expansion. To achieve this, we construct a regularization term to further guide the autoencoder-based network as:
\begin{equation}
\begin{small}
\mathcal{L}_A = \sum_{i=1}^{l} \sum_{j=1}^{m} {\bf 1}\left [ y_{i} = c_{j} \right ] \cdot \left [ 1 - \frac{S_{i}^{p+e}\cdot o_{j}}{\left \| S_{i}^{p+e} \right \|\left \| o_{j} \right \|}  \right ]\,
\end{small}
\end{equation}
where $S_{i}^{p+e}$ is the combined semantic feature representation of the $i$-th seen class example $x_{i}$, $y_{i}$ is the class label and $c_{j}$ is the $j$-th class label among $m$ classes. ${\bf 1}\left [ y_{i} = c_{j} \right ]$ is an indicator function that takes a value of one if its argument is true, and zero otherwise. Finally, combined with Eq. (2), the unified objective can be described as:
\begin{equation}
\small
\begin{aligned}
\mathcal{L} = &\alpha \cdot \underset{\mathcal{L}_{VAE}}{\underbrace{\sum_{x\sim \mathcal{D}}\left \| x - \hat{x} \right \|_{2}^{2}-D_{KL}  \big (q\left ( z | x \right )  \Vert p\left ( z \right ) \big )
}} \\
&+ \beta \cdot \underset{\mathcal{L}_A}{\underbrace{\sum_{i=1}^{l} \sum_{j=1}^{m} {\bf 1}\left [ y_{i} = c_{j} \right ] \cdot  \left [ 1 - \frac{S_{i}^{p+e}\cdot o_{j}}{\left \| S_{i}^{p+e} \right \|\left \| o_{j} \right \|}  \right ]}}\,,
\end{aligned}
\end{equation}
where $\mathcal{L}_{VAE}$ acts as a base term that mainly guides the reconstruction of the input visual examples. $\mathcal{L}_{A}$ is an alignment term that provides additional guidance to learning latent vectors and forces the manifold structure of the combined semantic feature space to approximate the structure of the embedded manifold extracted from the visual feature space. $\alpha$ and $\beta$ are two hyperparameters that control the balance between these two terms.

\subsubsection{Prototype Update}
After the expansion phase, we need to update the prototypes for each class. We have different strategies regarding the seen and unseen classes.

First, for each seen class. Because we obtained the trained autoencoder and all the seen class examples are available, we can simply compute the center, i.e., the mean value, of all latent vectors $z_{i}$ belonging to the same class and combine the center with the predefined prototype for each seen class as:
\begin{equation}
P^{e} = \frac{1}{h}\sum_{i=1}^{h}z_{i}\,,
\end{equation}
\begin{equation}
P = P^{p} \uplus P^{e}\,,
\end{equation}
where $z_{i}$ is the expanded semantic features obtained by the encoder, $h$ is the number of examples belonging to this specific seen class, $P^{p}$ and $P^{e}$ are predefined and expanded prototypes, respectively, and $\uplus$ denotes the operation that concatenates two vectors.

Second, for each unseen class, as no example is available during the whole training phase, we cannot apply the trained autoencoder to update the prototypes directly. Instead, we use another strategy by considering the local linearity among prototypes. Specifically, for each predefined unseen class prototype, we first obtain its $g$ nearest neighbors from predefined seen class prototypes. Then, we estimate each predefined unseen class prototype by a linear combination of its corresponding $g$ neighbors as:
\begin{equation}
\begin{aligned}
{P^{p}}' &= \theta _{1}P_{1}^{p} + \theta _{2}P_{2}^{p} + \cdots +\theta _{g}P_{g}^{p} \\
& = \theta P_{1\rightarrow g}^{p}\,,  
\end{aligned}
\end{equation}
where ${P^{p}}'$ is the predefined prototype of the unseen class, $\left \{ P_{i}^{p} \right \}_{i=1}^{g}$ are its $g$ nearest neighbors from predefined seen class prototypes, and $\left \{ \theta_{i} \right \}_{i=1}^{g}$ are the estimation parameters. Eq. (19) is a simple linear programming, and we can easily obtain the estimation parameters by solving a minimization problem as:
\begin{equation}
\theta = \underset{\theta}{\arg \min} \left \| {P^{p}}'- \theta P_{1\rightarrow g}^{p} \right \|.
\end{equation}
With the obtained estimation parameter $\theta$, we can update the prototype for each unseen class as:
\begin{equation}
\begin{aligned}
{P^{e}}' &= \theta _{1}P_{1}^{e} + \theta _{2}P_{2}^{e} + \cdots +\theta _{g}P_{g}^{e} \\
&= \theta P_{1\rightarrow g}^{e}\,,
\end{aligned} 
\end{equation}
\begin{equation}
{P}' = {P^{p}}' \uplus  {P^{e}}'\,,
\end{equation}
where ${P}'$ is the updated prototype for the unseen class and $\left \{ P_{i}^{e} \right \}_{i=1}^{g}$ are the corresponding $g$ expanded prototypes of its $g$ seen class neighbors.

\subsubsection{Recognition}
In our model, similar to some methods, we also adopt the simple autoencoder training framework \cite{kodirov2017semantic} to learn the projection between the visual and semantic feature spaces. The encoder $f_{e}(\cdot)$ first projects an example from the visual feature space to the semantic feature space to reach its prototype. Then, the decoder $f_{d}(\cdot)$ reversely projects it back to the visual feature space and reconstructs the example. The latent vectors of the autoencoder are forced to be the prototypes of each class. These two steps guarantee the robustness of our learned projection.

Our model mainly focuses on the alignment of manifold structures by semantic feature expansion, so we do not apply any additional technique to the projection training phase. A simple linear autoencoder with just one hidden layer is trained to obtain the visual-semantic projection. Specifically, the encoder $f_{e}(\cdot)$ can be regarded as the forward projection, and the decoder can be regarded as the reverse projection. In the testing phase, taking the forward projection as an example, we can project an unseen class example ${x_{i}}'$ to the semantic feature space to obtain its semantic feature representation $f_{e}({x_{i}}')$. As to the recognition, we simply search the most closely related prototype and set the class corresponding with this prototype to the testing example as:
\begin{equation}
\Omega  ({x_{i}}') = \underset{j}{arg min}Dist(f_{e}({x_{i}}'), {p_{j}}')
\end{equation}
where ${p_{j}}'$ is the prototype of the $j$-th unseen class, $Dist(\cdot, \cdot)$ is a distance measurement, and $\Omega(\cdot)$ returns the class label of the testing unseen class example.

\section{Experiment}

\subsection{Experimental Setup}

\subsubsection{Dataset}
Our model is evaluated on five widely used benchmark datasets for zero-shot learning, including Animals with Attributes (AWA) \cite{lampert2014attribute}, CUB-200-2011 Birds (CUB) \cite{wah2011caltech}, aPascal\&Yahoo (aPa\&Y) \cite{farhadi2009describing}, SUN Attribute (SUN) \cite{patterson2014sun}, and ILSVRC2012/ILSVRC2010 (ImageNet) \cite{russakovsky2015imagenet}. The first four are medium-scale datasets, and ImageNet is a large-scale dataset. The AWA \cite{lampert2014attribute} consists of 30,475 images of 50 animal classes, of which 40 are seen classes and the remaining 10 are unseen classes. Each class is represented by an 85-dimensional numeric attribute feature as the prototype. The 40 seen classes including 24,295 images are used for training, and the remaining 10 unseen classes with 6,180 images are used for testing. The CUB \cite{wah2011caltech} consists of 11,788 images of 200 bird species, from which 150 of them are seen classes and 50 are unseen classes. In this dataset, 8,855 images within 150 seen classes are used for training, and the remaining 2,933 images within 50 unseen classes are used for testing. Each of their prototypes is represented by a 312-dimensional semantic attribute feature. The aPa\&Y \cite{farhadi2009describing} consists of 15,339 images from the Pascal VOC 2008 and Yahoo. A 64-dimensional attribute feature is used as the prototype for each class. Among them, 20 classes with 12,695 images in the PASCAL VOC 2008 \cite{everingham2008pascal} train\&val set act as the seen classes, and 12 new classes including 2,644 images are used as the unseen classes. Each of the prototypes is represented by a 64-dimensional semantic attribute feature. The SUN \cite{patterson2014sun} consists of 14,340 scene images, from which two splits, 707/10 and 645/72, are commonly used. In our model, we only consider the latter split, which contains more unseen classes. A total of 645 seen classes are used for training, and the remaining 72 unseen classes are used for testing. Each of the prototypes is represented by a 102-dimensional semantic attribute feature. ImageNet \cite{russakovsky2015imagenet} consists of 1,360 classes, from which 1,000 from ILSVRC2012 are seen classes, and 360 from ILSVRC2010 are unseen classes. In the dataset, $2.0\times 10^5$ images within 1,000 seen classes are used for training, and the remaining $5.4\times 10^4$ images within 360 unseen classes are used for testing. Each of the prototypes is represented by a 1,000-dimensional semantic word vector. The basic description of these datasets is listed in TABLE I.

\begin{table}[t]
\renewcommand\thetable{I}
    \begin{center}
        \caption{Description of datasets}
        \setlength{\tabcolsep}{2.9mm}{  
            \begin{tabular}{ccccc}        
                \hline                   
                Dataset         & \# Examples       & \# SCs          & \# UCs            & D-SF \\
                \hline
                AWA             & 30475             & 40              & 10                & 85    \\ 
                CUB             & 11788             & 150             & 50                & 312   \\ 
                aPa\&Y          & 15339             & 20              & 12                & 64    \\
                SUN             & 14340             & 645             & 72                & 102    \\
                ImageNet        & $2.54\times 10^5$ & 1000            & 360               & 1000  \\
                \hline  
        \end{tabular}}
     \end{center}
     \label{tab1}  
\footnotesize{Notation: \# -- number, SCs/UCs -- seen/unseen classes, D-SF -- dimension of semantic feature.}
\end{table}

\subsubsection{Evaluation}
As a common practice in zero-shot learning, we use the Hit@k accuracy \cite{frome2013devise} to evaluate the performance of models. For each testing example, the model predicts its top-k possible class labels, and we correctly classify the example if and only if the ground truth label is within these predicted k class labels. We apply Hit@1 for AWA, CUB, aPa\&Y, and SUN, which is the normal accuracy. For ImageNet, we apply Hit@5 accuracy to fit a larger scenario following common practice.

In the experiments, we compare our proposed method with 18 competitors (TABLE III). The selection for the competitors is based on the following criteria: (1) all of these competitors are published in the most recent years; (2) they cover a wide range of models; (3) all of these competitors are under the same settings, i.e., datasets, evaluation criteria and the visual and semantic features adopted; and (4) they clearly represent the state-of-the-art. Moreover, as mentioned in Section II, our model and all selected competitors strictly comply with the nontransductive zero-shot setting that the training only relies on seen class examples, while the unseen class examples are only available during the testing phase.

\subsubsection{Implementation}
In our experiments, to consider all competitors, we also use GoogleNet \cite{szegedy2015going} to extract the visual features for image examples, from which each image is presented by a 1024-dimensional vector. Regarding the semantic features, we use the semantic attributes for AWA, CUB, aPa\&Y, and SUN and use the semantic word vectors for ImageNet. In our model, the autoencoder-based network for expansion and alignment has five layers. Specifically, the encoder and decoder parts contain two layers with (1024,256) and (256,1024) neurons, respectively. The central hidden layer represents the expected latent feature space, which can be adjusted to the dimension of semantic features we expand.
\begin{figure}[t]
    \centerline{\includegraphics[width=0.48\textwidth]{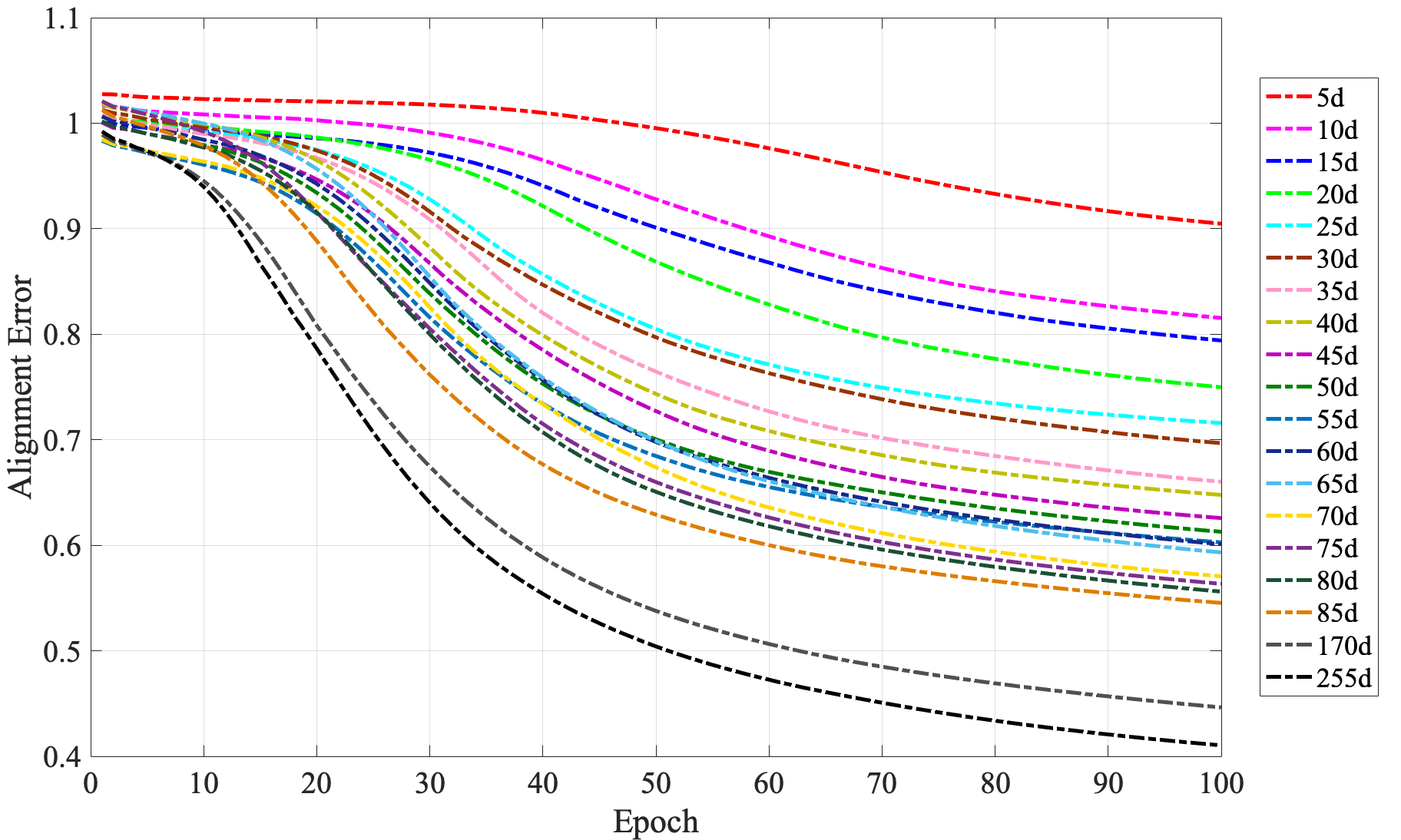}}
    \caption{Effects of expanded dimensions on alignment error. Curves in different colors represent the alignment error of different expanded dimensions ranging from 5 dimensions to 85 dimensions (expansion rate: 100\%) in an interval of 5 dimensions. In addition, we also show the results of 170 dimensions (expansion rate: 200\%) and 255 dimensions (expansion rate: 300\%) to better demonstrate the trends (better viewed in color).}
    \label{fig3}
\end{figure}
\begin{figure}[t]
    \centerline{\includegraphics[width=0.42\textwidth]{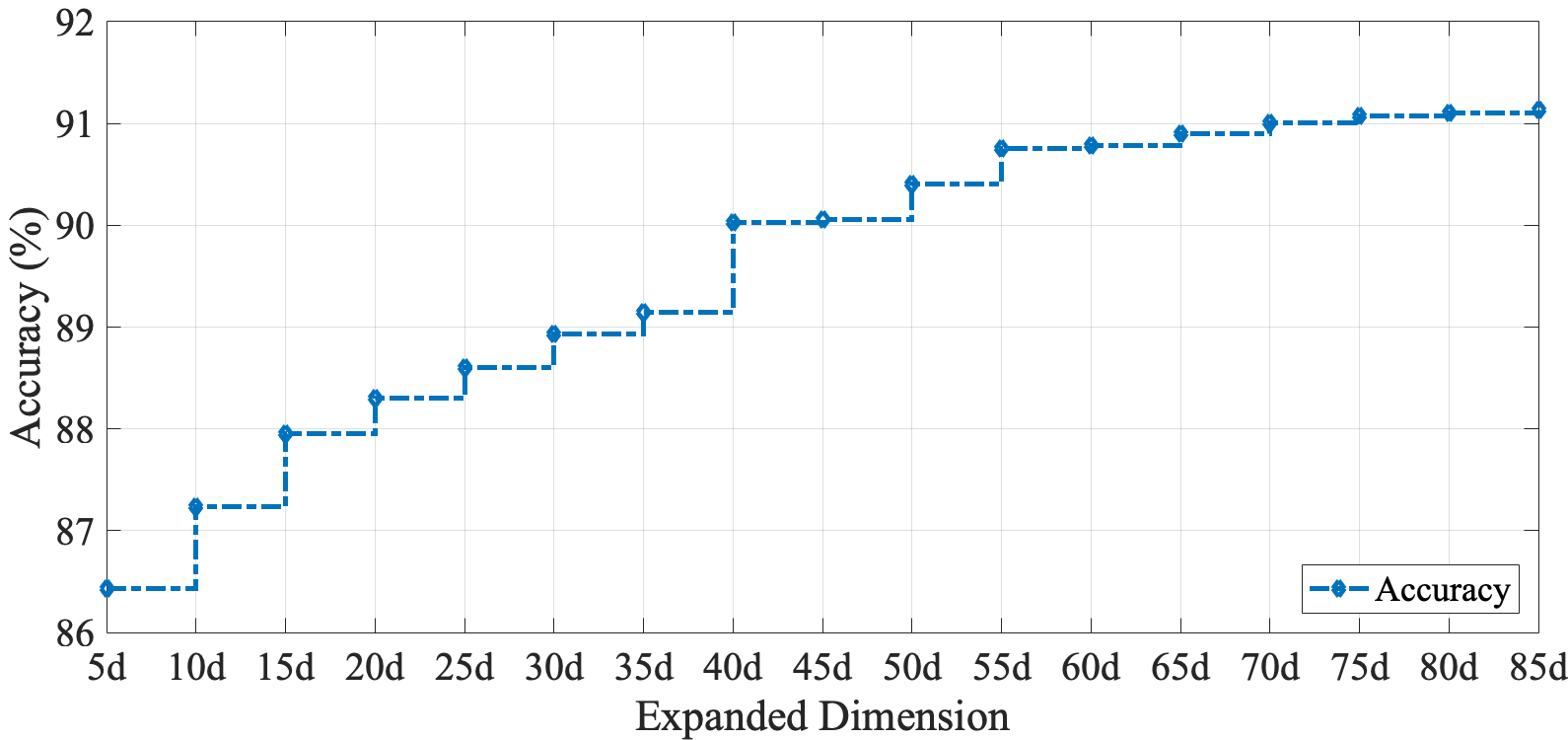}}
    \caption{Effects of expanded dimensions on prediction accuracy. The stepped curve represents the prediction accuracy under different expanded dimensions ranging from 5 dimensions to 85 dimensions (expansion rate: 100\%) in an interval of 5 dimensions.}
    \label{fig4}
\end{figure}
Theoretically, within a reasonable expansion range, we can expect that the more auxiliary semantic features we expand, the better the model performance will be. This is because with more auxiliary semantic features, more information can be utilized when training and inferring for the recognition. It will also be easier to align the manifold structures between the visual and semantic feature spaces; thus, the projection between these two feature spaces can be better trained and further mitigate the domain shift problem. Taking the AWA as an example, we conduct a comparison experiment to evaluate the effects of expanded dimensions of the auxiliary semantic features. The comparison results are shown in Figs. \ref{fig3} and \ref{fig4}. In Fig. \ref{fig3}, we can observe that within a reasonable expansion range, e.g., within an expansion rate of 100\% ($d_{expanded}/d_{pre-defined}=1$), as the expanded dimensions increase, the alignment error gradually decreases. In addition, we can also observe that even when the expansion rate reaches 200\% and 300\%, the trend of alignment error reduction remains. As to the prediction accuracy, it can be seen from Fig. \ref{fig4} that as the expanded dimensions increase, the accuracy also gradually increases and eventually becomes comparatively stable with larger expansion rates. Based on the above analysis, in our experiments, without loss of generality, we empirically apply a relatively medium expansion rate, i.e., $60\%\pm 15\%$, for all datasets except ImageNet. We expand 65, 138, 26, and 58 auxiliary semantic features for AWA, CUB, aPa\&Y, and SUN to fit the total semantic features as nice round numbers for 150, 450, 90, and 160. Regarding the ImageNet, because the dimensions of the visual features and predefined semantic features are 1,024 and 1,000, respectively, which results in expandable auxiliary semantic features ranging from 0 to 24 for a better projection. Thus, we fairly expand 12 auxiliary semantic features for ImageNet. In our model, the dimensional comparison of predefined and expanded semantic features for these five datasets is shown in TABLE II. As mentioned above, the neurons of the central hidden layer are adjusted to 65, 138, 26, 58, and 12 for AWA, CUB, aPa\&Y, SUN, and ImageNet, respectively.

\begin{figure}[h]
  \centering
  \subfigure[Reconstruction Loss]{
    \label{fig:subfig:a} 
    \includegraphics[width=3.0in]{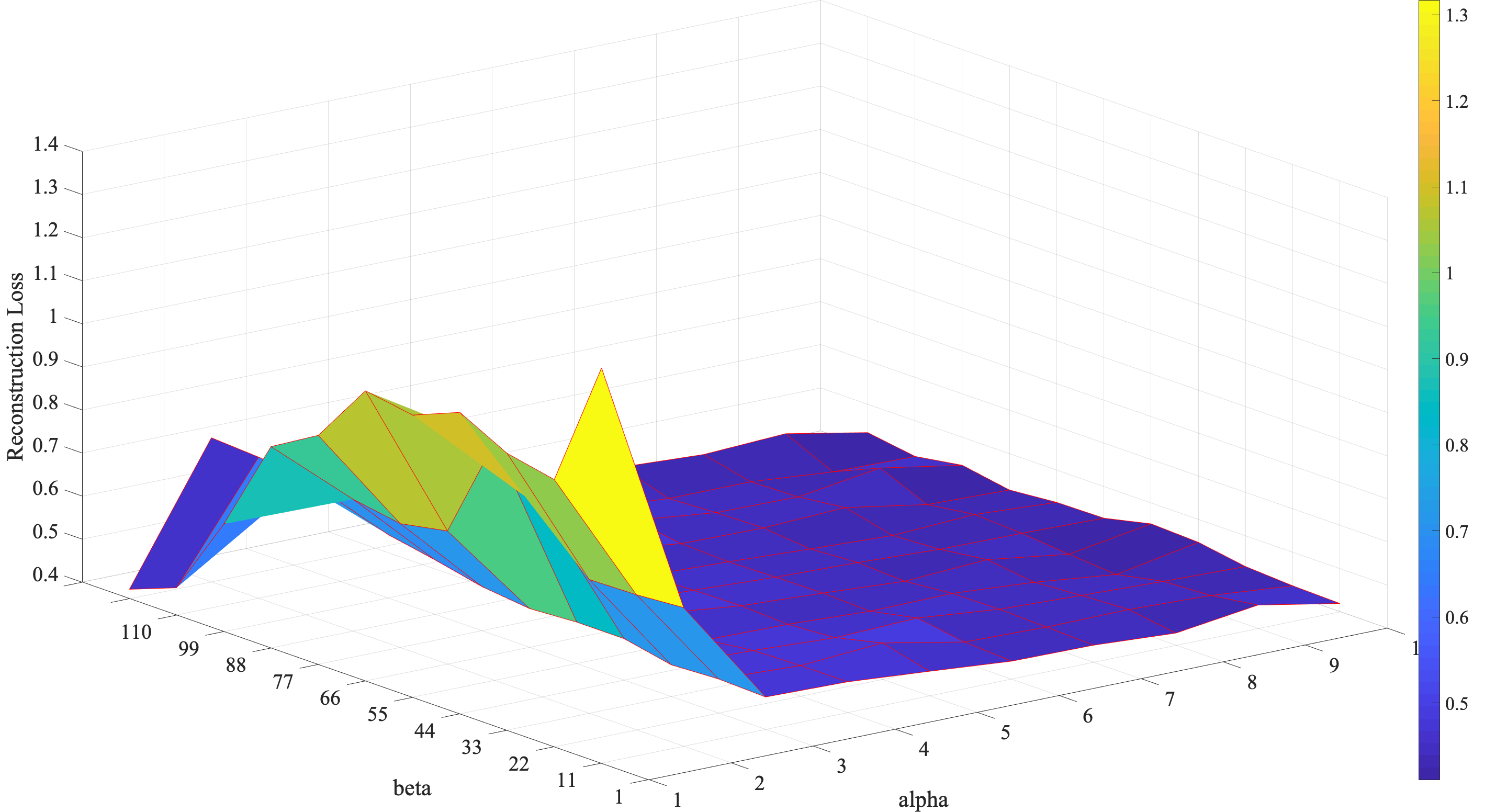}}
  \subfigure[Alignment Loss]{
    \label{fig:subfig:b} 
    \includegraphics[width=3.0in]{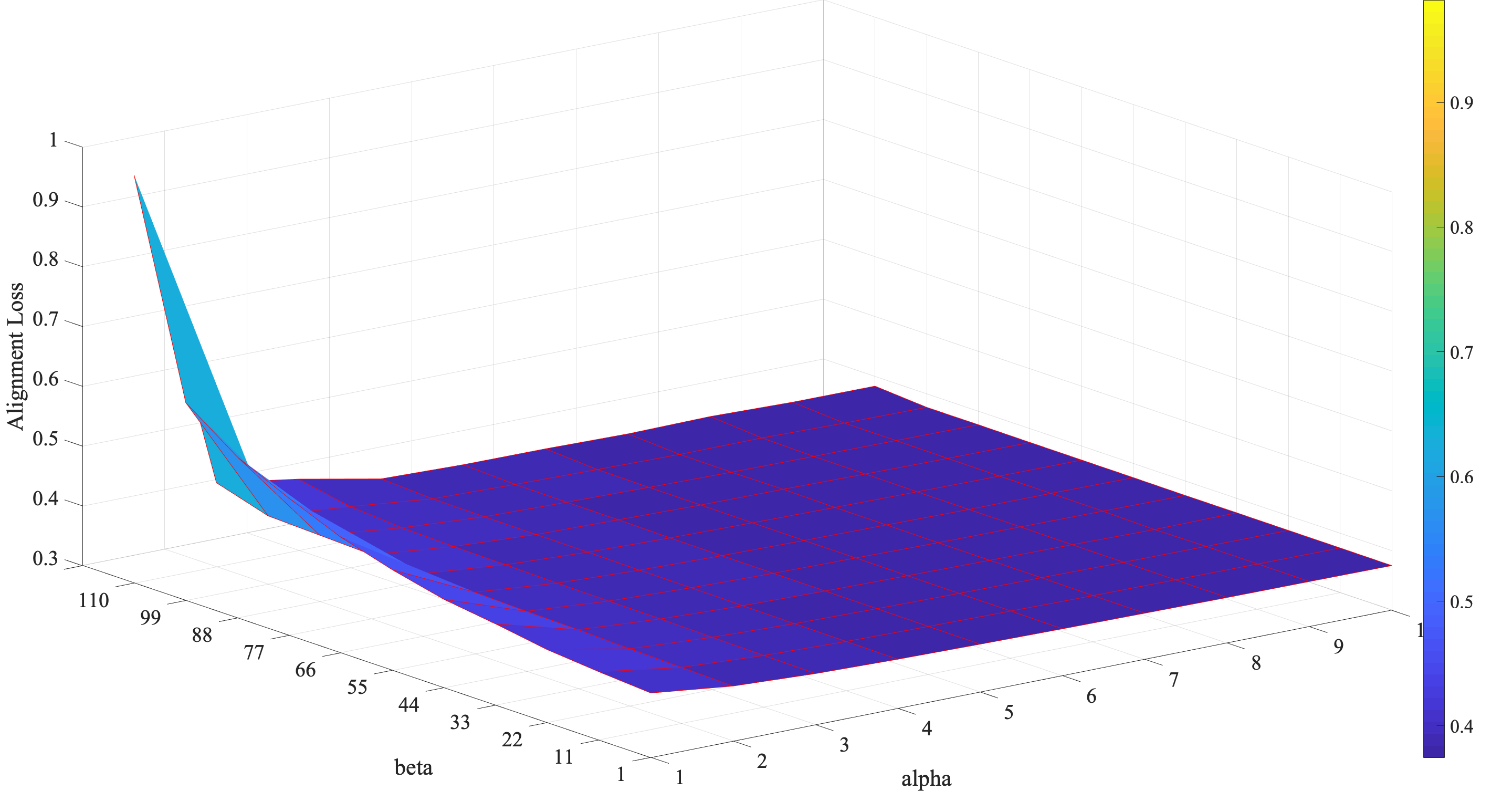}}
\subfigure[Total Loss]{
    \label{fig:subfig:c} 
    \includegraphics[width=3.0in]{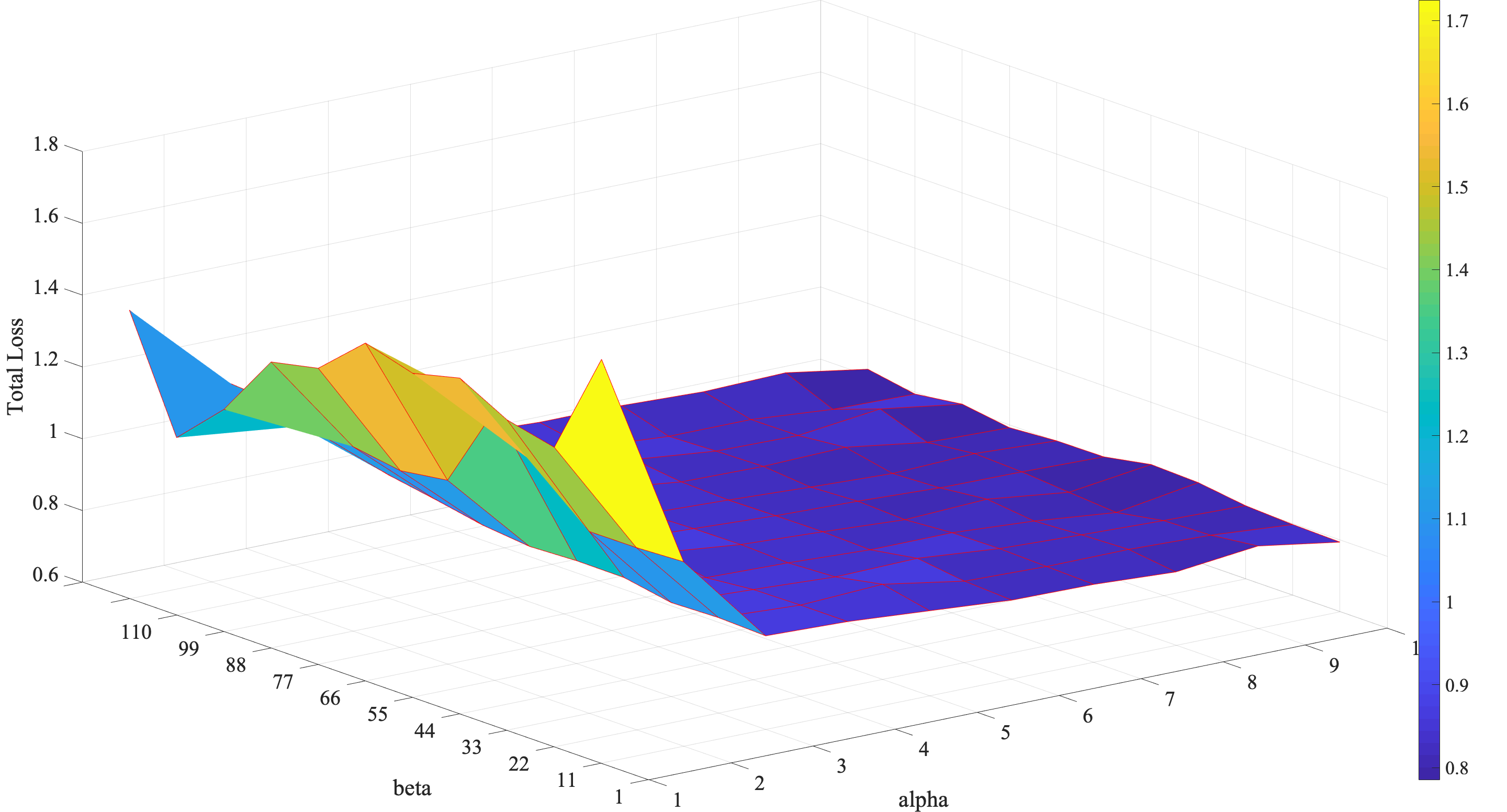}}
\caption{The sensibilities of hyperparameters $\alpha$ and $\beta$: (i) effects on reconstruction loss; (ii) effects on alignment loss; and (iii) effects on total loss (better viewed in color and zoom-in mode).}
  \label{fig5}
\end{figure}

For the hyperparameters $\alpha$ and $\beta$ in Eq. (16), we conduct a grid-search experiment \cite{hsu2003practical,guo2016improved} to evaluate their sensibilities. Based on the definition in Eq. (15), the alignment loss is normally considerably smaller than the reconstruction loss. Thus, a larger penalty, i.e., $\beta$, should be applied to the alignment term in Eq. (16) to accelerate the convergence of alignment loss \cite{kendall2018multi,sener2018multi}. In our grid-search experiment, we set the $\alpha$ and $\beta$ ranges within [1, 10] and [1, 110], respectively. The grid-search results are shown in Fig. \ref{fig5}, from which we can observe that the optimal $\alpha / \beta$ selections are 9/77 for the reconstruction loss, 10/110 for the alignment loss, and 9/77 for the total loss. Therefore, we choose to set the hyperparameters $\alpha$ and $\beta$ as 9 and 77, respectively. The hyperparameter $g$ is used to update the class prototypes of unseen classes. Based on our previous work \cite{guo2019adaptive} that adopted the same strategy by linearly associating with seen class prototypes, $g$ is not a critical parameter when ranged on a reasonable scale, e.g., several neighbors to tens of neighbors. Therefore, without loss of generality, we empirically choose to consider 8 nearest neighbors in our model. As to the recognition phase, a simple linear autoencoder with just one hidden layer is trained to obtain the visual-semantic projection, i.e., the neurons of input/output are equal to the visual features, and the neurons of the central hidden layer are equal to the semantic features. The cosine similarity is applied to search the related prototype of the testing examples.

\begin{table*}[htbp]
    \renewcommand\thetable{III}
    \centering
    \setlength{\tabcolsep}{3.9mm}{
    \begin{threeparttable}  
        \caption{Comparison with state-of-the-art competitors}  
        \label{table2}    
        \begin{tabular}{lcccccccccc}  
            \toprule  
            \multirow{2}{*}{Method}&  
            \multicolumn{2}{c}{AWA}&\multicolumn{2}{c}{CUB}&\multicolumn{2}{c}{aPa\&Y}&\multicolumn{2}{c}{SUN}&\multicolumn{2}{c}{ImageNet}\cr  
            \cmidrule(lr){2-3} \cmidrule(lr){4-5} \cmidrule(lr){6-7} \cmidrule(lr){8-9} \cmidrule(lr){10-11} 
            &SS  &ACC               &SS   &ACC              &SS   &ACC                 &SS   &ACC              &SS   &ACC\cr  
            \midrule  
            DeViSE \cite{frome2013devise}            &A/W &56.7/50.4         &A/W  &33.5             &-    &-                   &-    &-                &A/W  &12.8\cr
            DAP \cite{lampert2014attribute}               &A   &60.1              &A    &-                &A    &38.2                &A    &72.0             &-    &-\cr
            MTMDL \cite{yang2014unified}             &A/W &63.7/55.3         &A/W  &32.3             &-    &-                   &-    &-                &-    &-\cr
            ESZSL \cite{romera2015embarrassingly}             &A   &75.3              &A    &48.7             &A    &24.3                &A    &82.1             &-    &-\cr
            SSE \cite{zhang2015zero}               &A   &76.3              &A    &30.4             &A    &46.2                &A    &82.5             &-    &-\cr
            RRZSL \cite{shigeto2015ridge}             &A   &80.4              &A    &52.4             &A    &48.8                &A    &84.5             &W    &-\cr
            Ba et al. \cite{ba2015predicting}         &A/W &69.3/58.7         &A/W  &34.0             &-    &-                   &-    &-                &-    &-\cr    
            AMP \cite{bucher2016improving}               &A+W &66.0              &A+W  &-                &-    &-                   &-    &-                &A+W  &13.1\cr
            JLSE \cite{zhang2016zero}              &A   &80.5              &A    &41.8             &A    &50.4                &A    &83.8             &-    &-\cr
            SynC$^{struct}$ \cite{changpinyo2016synthesized}   &A   &72.9              &A    &54.4             &-    &-                   &-    &-                &-    &-\cr
            MLZSC \cite{bucher2016improving}             &A   &77.3              &A    &43.3             &-    &53.2                &A    &84.4             &-    &-\cr
            SS-voc \cite{fu2016semi}            &A/W &78.3/68.9         &A/W  &-                &-    &-                   &-    &-                &A/W  &16.8\cr
            SAE \cite{kodirov2017semantic}              &A   &84.7              &A    &61.2             &A    &55.1                &A    &91.0             &W    &26.3\cr     
            CLN+KRR \cite{long2017zero}           &A   &81.0              &A    &58.6             &-    &-                   &-    &-                &-    &-\cr
            MFMR \cite{xu2017matrix}              &A   &76.6              &A    &46.2             &A    &46.4                &A    &81.5             &-    &-\cr
            RELATION NET \cite{yang2018learning}     &A   &84.5              &A    &62.0             &-    &-                   &-    &-                &-    &-\cr
            CAPD-ZSL \cite{rahman2018unified}          &A   &80.8              &A    &45.3             &A    &55.0                &A    &87.0             &W    &23.6\cr
            LSE \cite{yu2018zero}               &A   &81.6              &A    &53.2             &A    &53.9                &-    &-                &W    &\underline{27.4}\cr
            AMS-SFE (AE, Ours)           &A   &\underline{\bf 90.9}        &A    &{\bf 67.8}       &A    &{\bf 59.4}          &A    &{\bf 92.7}       &W    &{\bf 26.1}\cr  
            AMS-SFE (VAE, Ours)           &A   &{\bf 90.2}        &A    &\underline{\bf 70.1}       &A    &\underline{\bf 59.7}          &A    &\underline{\bf 92.9}       &W    &{\bf 26.3}\cr  
            \bottomrule  
        \end{tabular}
        \label{tab3}  
        \footnotesize{SS, A, and W represents semantic space, attribute and word vectors, respectively; '/' represents 'or' and '+' represents 'and'; '-' represents that there is no reported result. ACC stands for accuracy (\%), where Hit@1 is used for AWA, CUB, aPa\&Y, and SUN, and Hit@5 is used for ImageNet. The results of our model are shown in bold, and the underline denotes the best result among all competitors.}
    \end{threeparttable}}
\end{table*} 

\begin{table}[t]
\renewcommand\thetable{II}
    \begin{center}
        \caption{Dimension of predefined (P) / expanded (E) features}
        \setlength{\tabcolsep}{1.8mm}{  
            \begin{tabular}{cccccc}        
                \hline                   
                           & AWA     & CUB     & aPa\&Y  & SUN     & ImageNet\\
                \hline
                P                & 85      & 312     & 64      & 102     & 1000 \\
                E                & 65      & 138     & 26      & 58      & 12 \\
                P+E              & 150     & 450     & 90      & 160     & 1012 \\ 
                \hline  
        \end{tabular}}
    \end{center}
    \label{tab2}    
\end{table}

\subsection{Results and Analysis}

\subsubsection{Compare with State-of-the-Art}
The comparison results with the selected state-of-the-art competitors are shown in TABLE III. It can be seen from the results that our model outperforms all competitors with great advantages in four datasets, including AWA, CUB, aPa\&Y, and SUN. The prediction accuracy of our model achieves 90.9\%, 70.1\%, 59.7\%, and 92.9\%, respectively. In ImageNet, our model obtains suboptimal performance among all competitors. Our prediction accuracy is 26.3\%, which is slightly weaker than LSE \cite{yu2018zero} and similar to SAE \cite{kodirov2017semantic}. It should be noted that these similar results in ImageNet of SAE and our model may be caused by the following two reasons. First, as mentioned in Sections III.B.5 and IV.A.3, our model also adopts a simple autoencoder training framework to obtain the projection between the visual and semantic feature spaces, which is similar to some existing methods such as SAE. Second, it can be seen in TABLE II that there are only 12 expanded semantic features for ImageNet, which also makes the total semantic features similar to the predefined features. In TABLE I, we can observe that the dimension of the predefined semantic features of ImageNet is 1,000. However, the dimension of the visual features is 1,024. Based on the analysis in Section IV.A.3, we only expand 12 auxiliary semantic features for ImageNet, which are far fewer than the predefined features. This limitation makes it difficult to perform the alignment between the visual and semantic feature spaces for ImageNet. Thus, the improvement regarding ImageNet is not that significant. In contrast, we have enough space to expand the auxiliary semantic features for AWA, CUB, aPa\&Y, and SUN, so the alignment can be better approximated for them. In our model, the dimensions of the expanded auxiliary semantic features for these , datasets are 65, 138, 26, 58, and 12, respectively.

From the comparison results, we can also observe that the performance of our VAE-based model is slightly better than the AE-based model for CUB, aPa\&Y, SUN, and ImageNet, from which the most significant improvement is from 67.8\% to 70.1\% for CUB. One possible reason is that CUB is also a good benchmark dataset for fine-grained image recognition tasks, except for ZSL, which consists of multiple bird species, and the transition between each class is relatively smoother and more continuous. Therefore, it is more likely to maintain these features in the semantic feature space, which helps the prediction by using the VAE.

\begin{figure*}[t]
  \centering
  \subfigure[SAE for AWA]{
    \label{fig:subfig:a} 
    \includegraphics[width=2.3in]{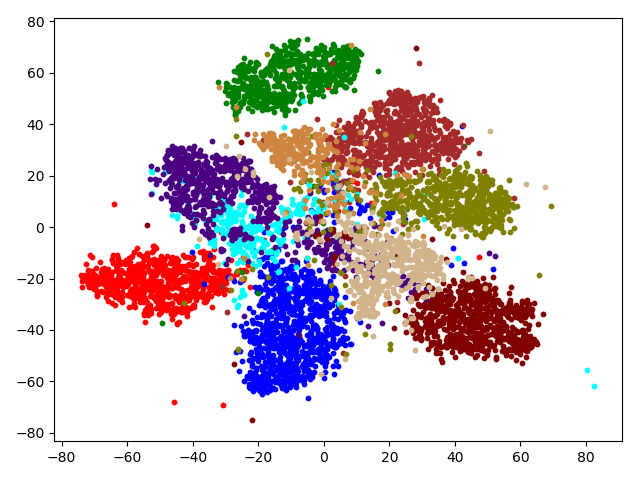}}
  \subfigure[AMS-SFE(AE) for AWA]{
    \label{fig:subfig:b} 
    \includegraphics[width=2.3in]{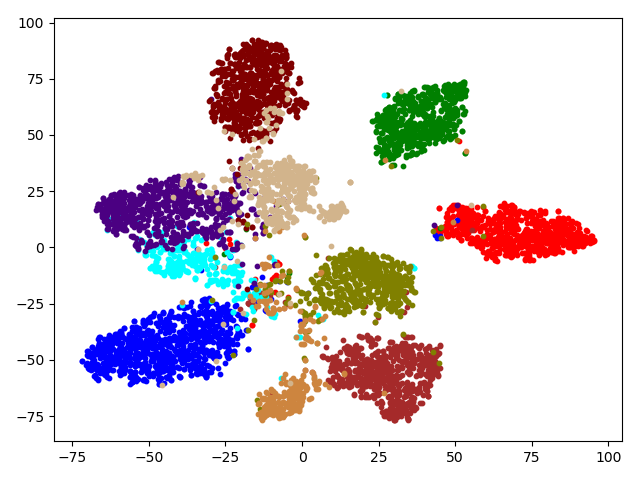}}
\subfigure[AMS-SFE(VAE) for AWA]{
    \label{fig:subfig:c} 
    \includegraphics[width=2.3in]{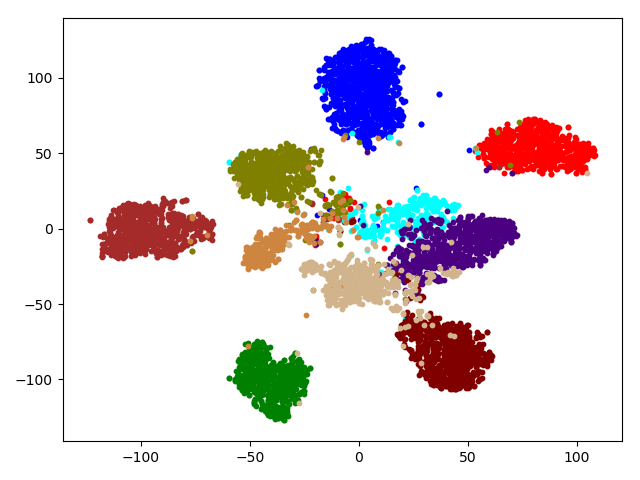}}

  \subfigure[SAE for CUB]{
    \label{fig:subfig:d} 
    \includegraphics[width=2.3in]{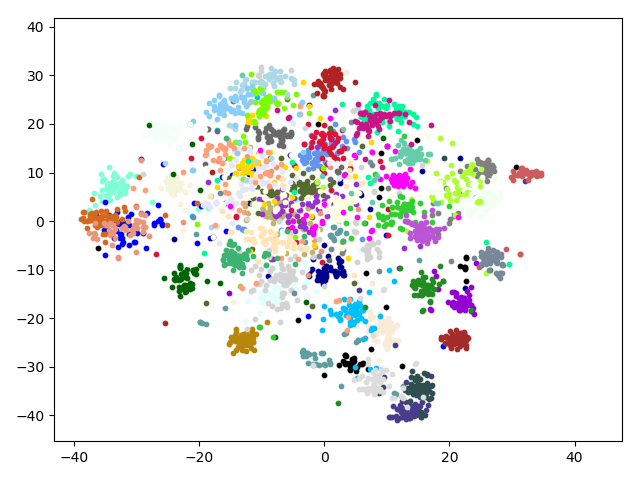}}
  \subfigure[AMS-SFE(AE) for CUB]{
    \label{fig:subfig:e} 
    \includegraphics[width=2.3in]{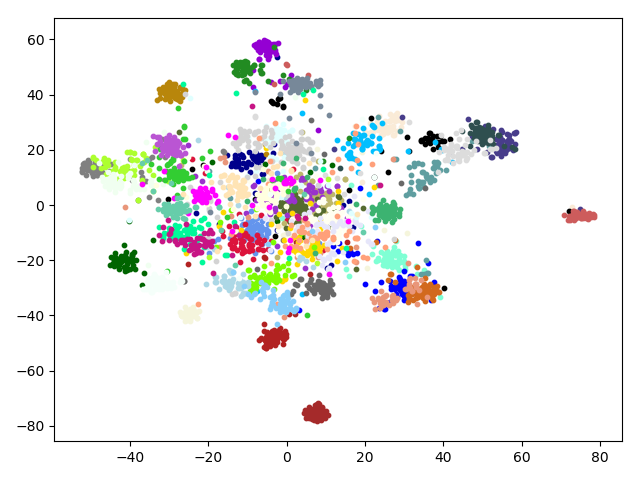}}
\subfigure[AMS-SFE(VAE) for CUB]{
    \label{fig:subfig:f} 
    \includegraphics[width=2.3in]{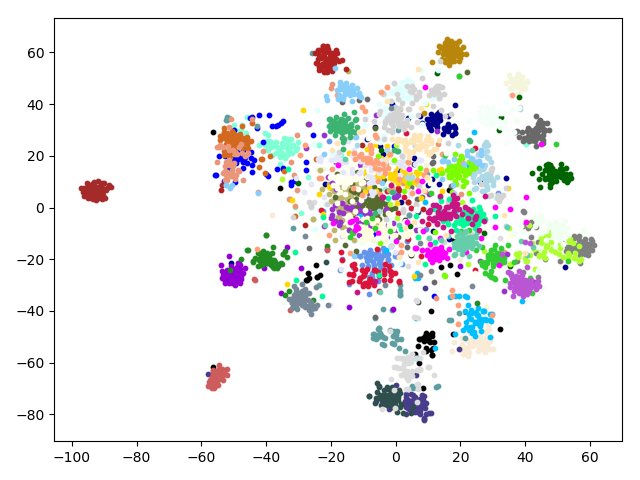}}
\caption{Visualization results of our model and the strongest competitor SAE. The upper part is the results for AWA, and the lower part is the results for CUB. (i) and (iv) are the visualization of SAE, (ii) and (v) are the visualization of our AE-based model, and (iii) and (vi) are the visualization of our VAE-based model (better viewed in color).}
  \label{fig6}
\end{figure*}

\begin{table}[h]
\renewcommand\thetable{IV}
    \begin{center}
        \caption{Hit@k accuracy (\%) for AWA, $k\in \left [ 1,5 \right ]$}
        \setlength{\tabcolsep}{0.9mm}{  
            \begin{tabular}{cccccc}        
                \hline                   
                Method           & Hit@1   & Hit@2   & Hit@3   & Hit@4   & Hit@5\\
                \hline
                SAE              & 84.7    & 93.5    & 97.2    & 98.8    & 99.4 \\
                AMS-SFE (ours)   & 90.9    & 97.4    & 99.5    & 99.8    & 99.8 \\ 
                \hline  
        \end{tabular}}
    \end{center}
   \label{tab4}   
\end{table}

\begin{figure}[h]
    \centerline{\includegraphics[width=0.37\textwidth]{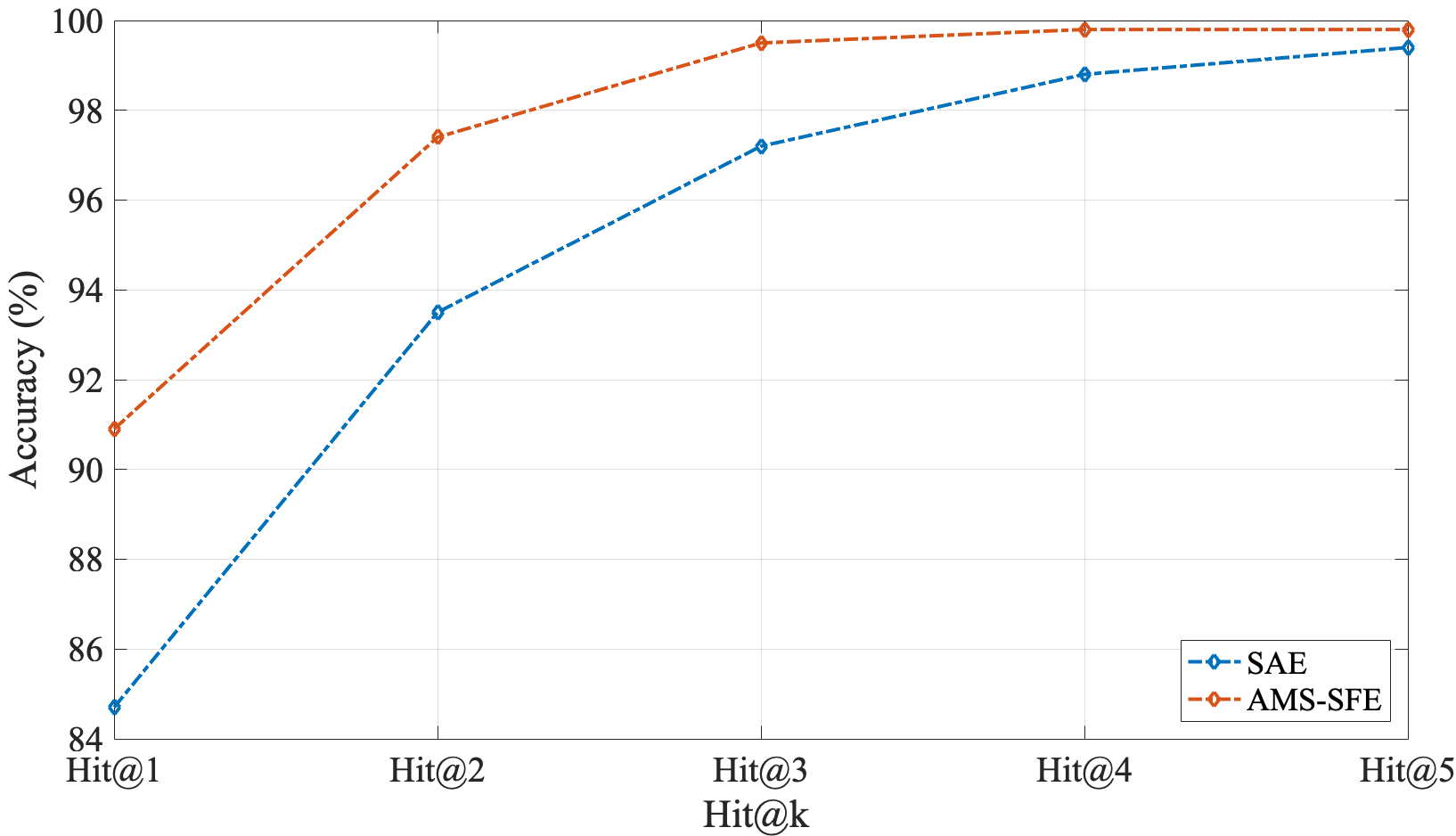}}
    \caption{Comparison of Hit@k accuracy for AWA. The blue curve denotes the SAE and the red curve denotes our model.}
    \label{fig7}
\end{figure}

\subsubsection{Projection Robustness}
We conduct the evaluation of projection robustness on AWA and CUB. AWA consists of 10 unseen classes, and CUB consists of 50 unseen classes. Our model is compared with the overall strongest competitor SAE \cite{kodirov2017semantic} to verify the projection robustness. A projection that maps from the visual to the semantic feature spaces is trained on seen class examples with our model. Then, we apply the projection to all unseen class examples to obtain their semantic features and visualize them in a 2D map by t-SNE \cite{maaten2008visualizing}. The visualization results are shown in Fig. \ref{fig6}. The upper part is the results for AWA, and the lower part is the results for CUB. In AWA, we observe that by using our model, only a small portion of these unseen class examples are misprojected in the semantic feature space. Moreover, due to the alignment of manifold structures between the visual and semantic feature spaces, these misprojected examples are less shifted from their class prototypes, which means that our model can also obtain better results for Hit@k accuracy when k varies and converges faster to the best performance. The comparison is shown in TABLE IV and Fig. \ref{fig7}. In CUB, the class number is much larger than that of AWA, so the visualization results seem more complicated and intertwined in the 2D map, but we still observe that our model can obtain more continuous and smoother semantic features in a wider visual field.

\begin{figure*}[t]
  \centering
  \subfigure[SAE for AWA]{
    \label{fig:subfig:a} 
    \includegraphics[width=2.8in]{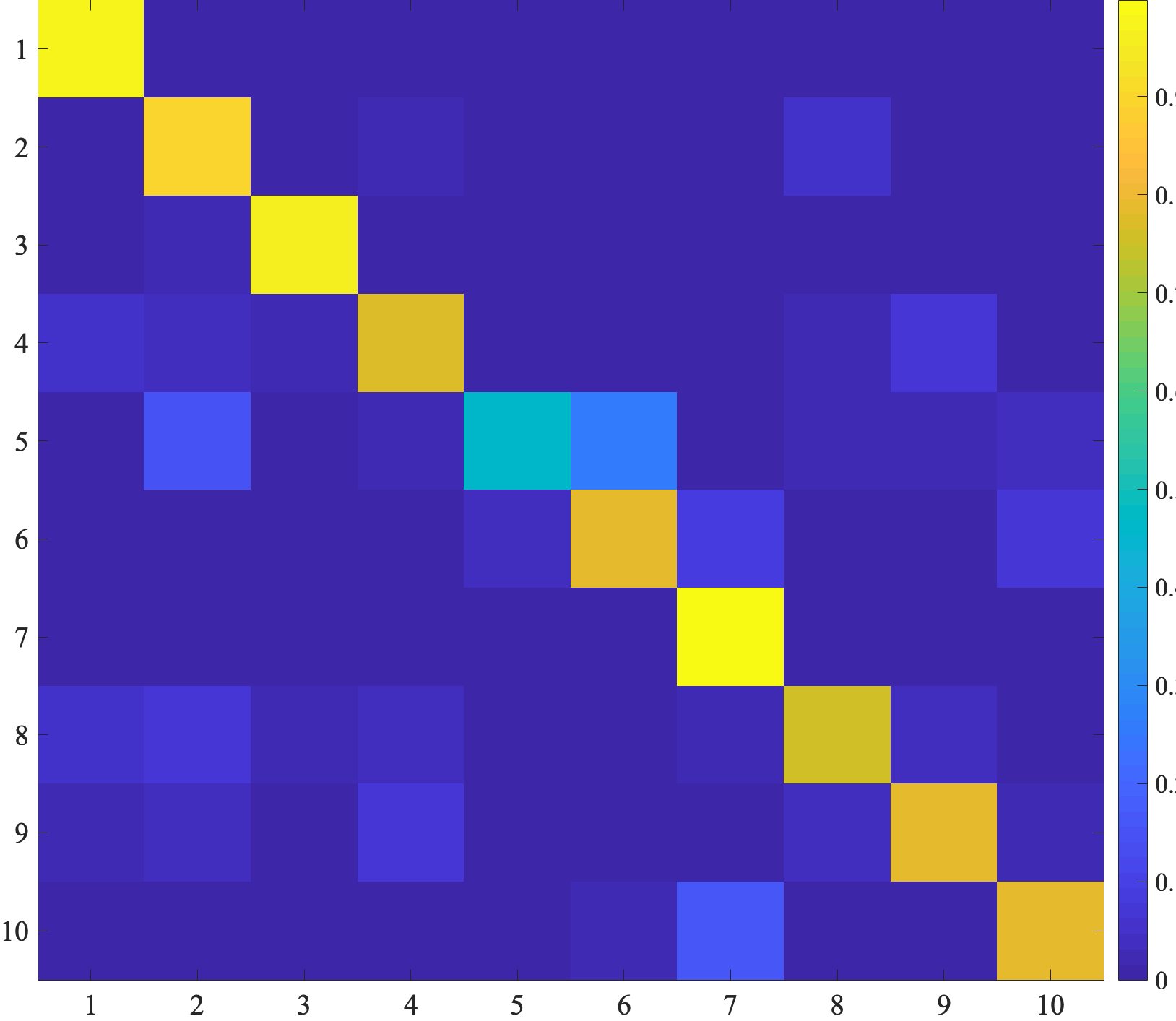}}
  \hspace{.47in}
\subfigure[AMS-SFE for AWA]{
    \label{fig:subfig:b} 
    \includegraphics[width=2.8in]{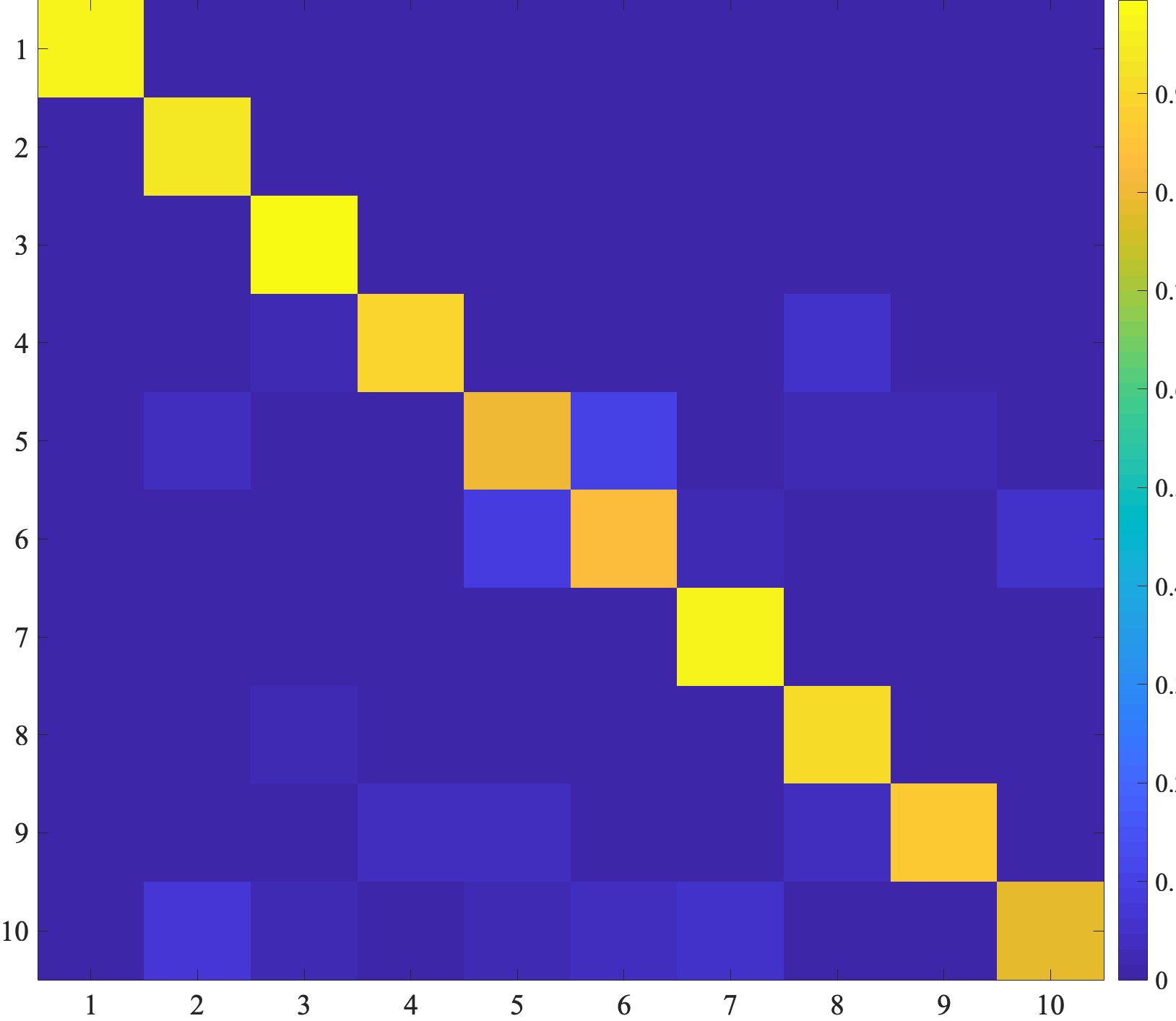}}
  \subfigure[SAE for CUB]{
    \label{fig:subfig:c} 
    \includegraphics[width=2.8in]{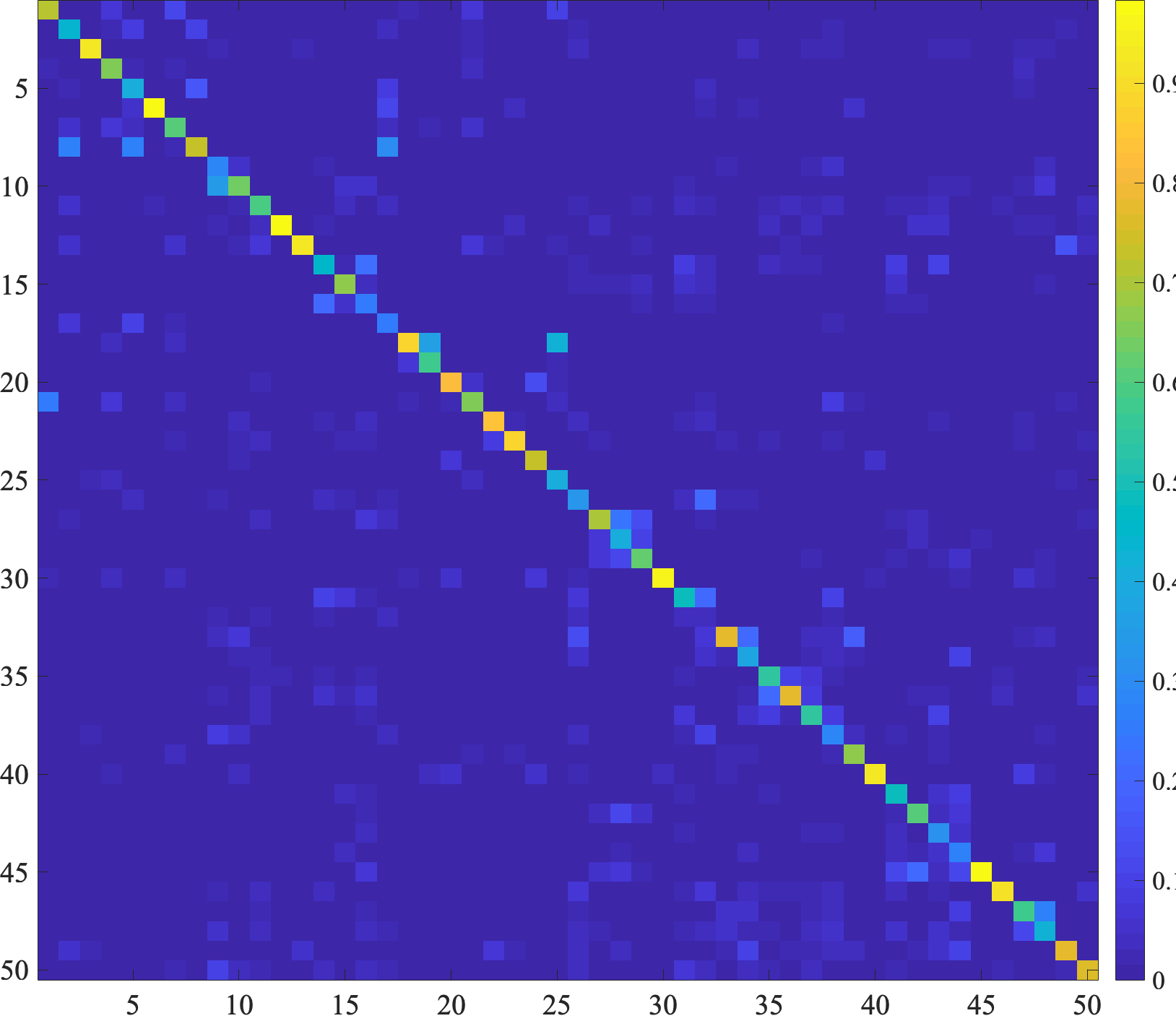}}
  \hspace{.47in}
\subfigure[AMS-SFE for CUB]{
    \label{fig:subfig:d} 
    \includegraphics[width=2.8in]{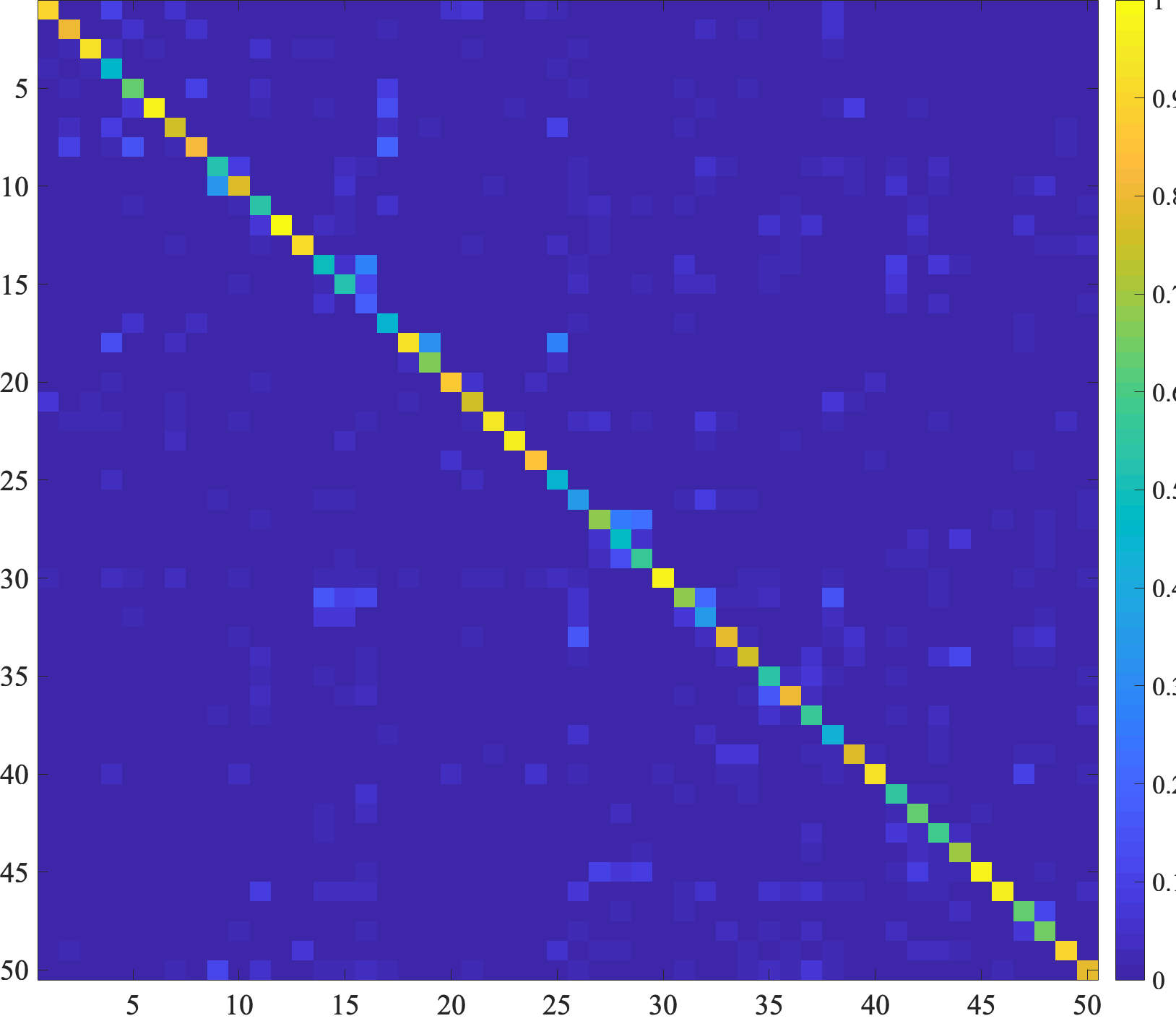}}
\caption{The confusion matrix on AWA and CUB. The upper part is the results for AWA, and the lower part is the results for CUB. (i) and (iii) are the confusion matrixes of SAE; (ii) and (iv) are the confusion matrixes of our model. The diagonal position indicates the classification accuracy for each class, the column position means the ground truth, and the row position denotes the predicted results (better viewed in color).}
  \label{fig8} 
\end{figure*}

\subsubsection{Fine-Grained Accuracy}
To further evaluate the predictive power of our model, we record and count the prediction results for each unseen class example and analyze the per-class performance. This evaluation is conducted on AWA and CUB, and we also compare our model with the overall strongest competitor SAE \cite{kodirov2017semantic}. The results are presented by the confusion matrixes in Fig. \ref{fig8}. The upper part is the results for AWA, and the lower part is the results for CUB. In each confusion matrix, the diagonal position indicates the classification accuracy for each class. The column position indicates the ground truth, and the row position denotes the predicted results. It can be seen from the results that our model can obtain higher accuracy, along with a more balanced and robust prediction for each unseen class for both AWA and CUB.

\begin{table}[h]
\renewcommand\thetable{V}
    \begin{center}
        \caption{Ablation comparison (accuracy\%) on predefined (P) / expanded (E) semantic features and both (P+E)}
        \setlength{\tabcolsep}{2.2mm}{  
            \begin{tabular}{cccccc}        
                \hline                   
                           & AWA     & CUB     & aPa\&Y  & SUN     & ImageNet\\
                \hline
                P                & 84.4    & 60.3    & 53.1    & 88.7  & 26.1 \\
                E                & 75.2    & 52.8    & 45.5    & 77.4    & 14.2 \\
                P+E              & 90.9    & 67.8    & 59.4    & 92.7    & 26.1 \\ 
                \hline  
        \end{tabular}}
    \end{center} 
    \label{tab5}  
\end{table}

\begin{figure}[h] 
    \centerline{\includegraphics[width=0.43\textwidth]{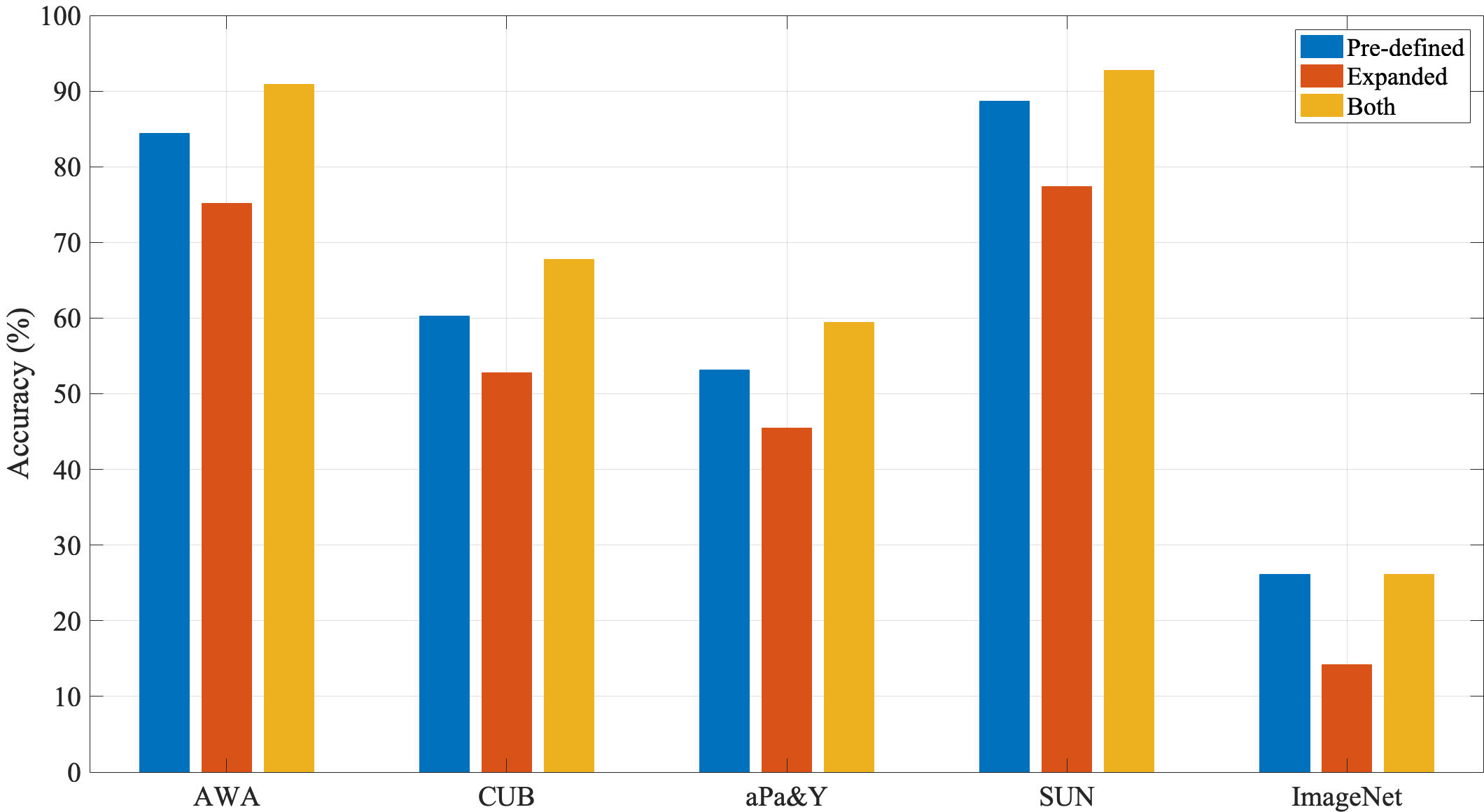}}
    \caption{Ablation comparison on five benchmark datasets. The blue bar denotes the predefined semantic feature, the yellow bar denotes the expanded semantic features, and the red bar denotes both.}
    \label{fig9}
\end{figure}

\subsubsection{Ablation Comparison}
We conduct an ablation experiment to further evaluate the effectiveness of our model. The performance is compared on all five benchmark datasets, including AWA, CUB, aPa\&Y, SUN, and ImageNet, with our AE-based model on three scenarios: (1) only predefined semantic features are used; (2) only expanded auxiliary semantic features are used; and (3) both predefined and expanded auxiliary semantic features are used. The comparison results are shown in TABLE V and Fig. \ref{fig9}. It can be seen from the results that our model greatly improves the performance of zero-shot learning by performing the alignment with the expanded auxiliary semantic features. Nevertheless, it should also be noted that, due to the very few (i.e., 12) expanded auxiliary semantic features for ImageNet as analyzed in Section IV.A.3, the model obtains similar performance as 26.1\% for both scenarios (1) and (3), where only the predefined semantic features and both semantic features are used, respectively. Thus, the impact of expanded semantic features on ImageNet is limited in our model, and the visual and semantic feature spaces cannot be better aligned. We may further investigate this problem in our future work.

\begin{figure*}[htbp]
  \centering
  \subfigure[AWA]{
    \label{fig:subfig:a} 
    \includegraphics[width=2.0in]{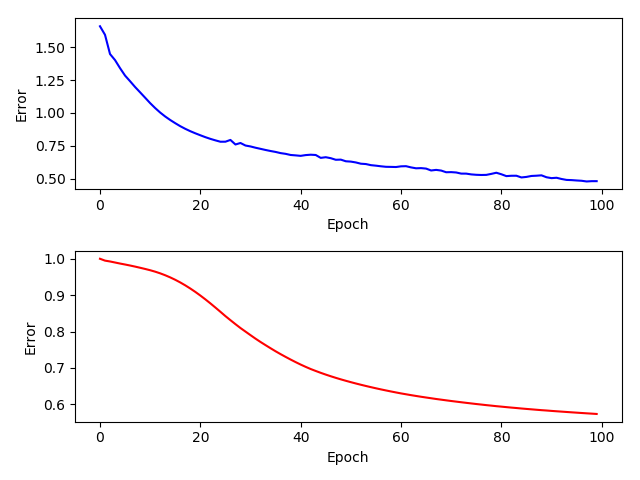}}
  \subfigure[CUB]{
    \label{fig:subfig:b} 
    \includegraphics[width=2.0in]{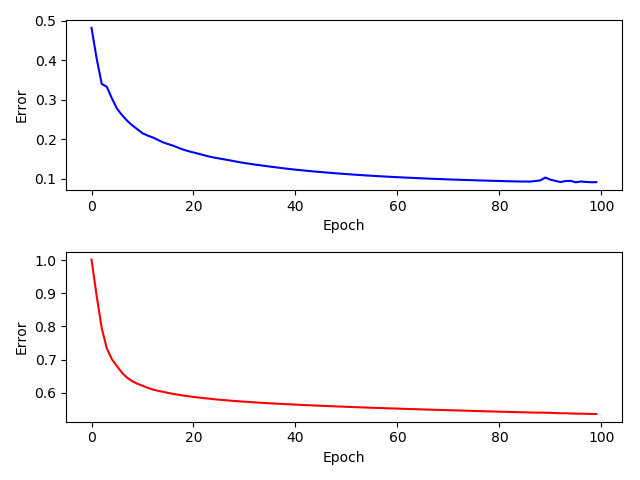}}
\subfigure[aPa\&Y]{
    \label{fig:subfig:c} 
    \includegraphics[width=2.0in]{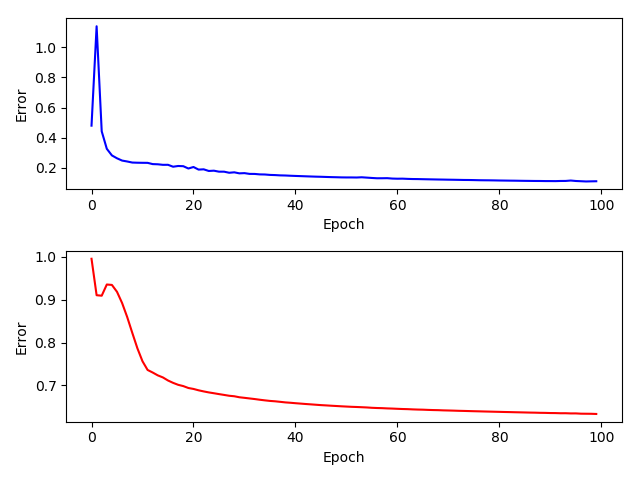}}
\subfigure[SUN]{
    \label{fig:subfig:d} 
    \includegraphics[width=2.0in]{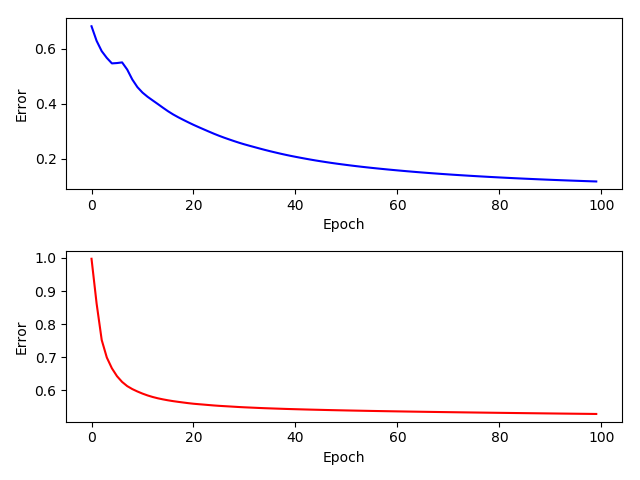}}
\subfigure[ImageNet]{
    \label{fig:subfig:e} 
    \includegraphics[width=2.0in]{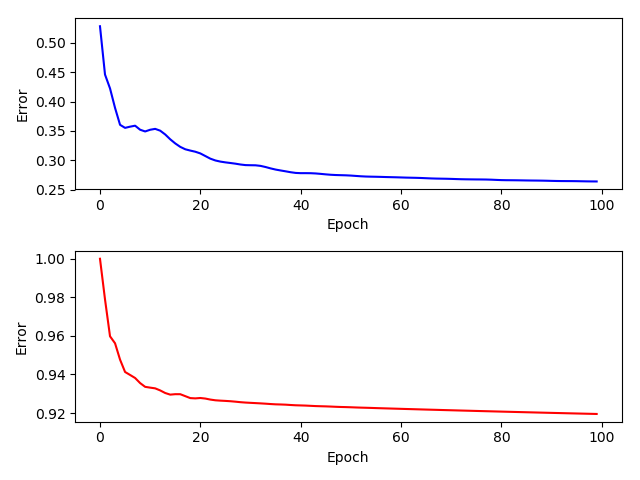}}
\caption{The convergence curves of AWA, CUB, aPa\&Y, SUN, and ImageNet. The blue curve denotes the reconstruction error, and the red curve denotes the alignment error.}
  \label{fig10} 
\end{figure*}

\section{Conclusion}
In this paper, we proposed a novel model called AMS-SFE for zero-shot learning. Our model aligns the manifold structures between the visual and semantic feature spaces by jointly conducting semantic feature expansion. Our model can better mitigate the domain shift problem and obtain a more robust and generalized visual-semantic projection. In the future, we have two research routes to further improve zero-shot learning. The first route investigates a more efficient and generalized method to further empower the semantic feature space, especially for some more challenging datasets such as ImageNet, whose semantic feature space is mainly based on the word vectors. The second route goes from the coarse-grained model to the fine-grained model to better model the subtle differences among different classes.

\ifCLASSOPTIONcaptionsoff
  \newpage
\fi

\bibliographystyle{IEEEtran}
\bibliography{my}

\begin{thebibliography}{10}
\providecommand{\url}[1]{#1}
\csname url@samestyle\endcsname
\providecommand{\newblock}{\relax}
\providecommand{\bibinfo}[2]{#2}
\providecommand{\BIBentrySTDinterwordspacing}{\spaceskip=0pt\relax}
\providecommand{\BIBentryALTinterwordstretchfactor}{4}
\providecommand{\BIBentryALTinterwordspacing}{\spaceskip=\fontdimen2\font plus
\BIBentryALTinterwordstretchfactor\fontdimen3\font minus
  \fontdimen4\font\relax}
\providecommand{\BIBforeignlanguage}[2]{{%
\expandafter\ifx\csname l@#1\endcsname\relax
\typeout{** WARNING: IEEEtran.bst: No hyphenation pattern has been}%
\typeout{** loaded for the language `#1'. Using the pattern for}%
\typeout{** the default language instead.}%
\else
\language=\csname l@#1\endcsname
\fi
#2}}
\providecommand{\BIBdecl}{\relax}
\BIBdecl

\bibitem{krizhevsky2012imagenet}
A.~Krizhevsky, I.~Sutskever, and G.~E. Hinton, ``Imagenet classification with
  deep convolutional neural networks,'' in \emph{Advances in neural information
  processing systems}, 2012, pp. 1097--1105.

\bibitem{zhao2017diversified}
B.~Zhao, X.~Wu, J.~Feng, Q.~Peng, and S.~Yan, ``Diversified visual attention
  networks for fine-grained object classification,'' \emph{IEEE Transactions on
  Multimedia}, vol.~19, no.~6, pp. 1245--1256, 2017.

\bibitem{zhang2018multilabel}
J.~Zhang, Q.~Wu, C.~Shen, J.~Zhang, and J.~Lu, ``Multilabel image
  classification with regional latent semantic dependencies,'' \emph{IEEE
  Transactions on Multimedia}, vol.~20, no.~10, pp. 2801--2813, 2018.

\bibitem{hu2014discriminative}
J.~Hu, J.~Lu, and Y.-P. Tan, ``Discriminative deep metric learning for face
  verification in the wild,'' in \emph{Proceedings of the IEEE conference on
  computer vision and pattern recognition}, 2014, pp. 1875--1882.

\bibitem{sun2014deep}
Y.~Sun, Y.~Chen, X.~Wang, and X.~Tang, ``Deep learning face representation by
  joint identification-verification,'' in \emph{Advances in neural information
  processing systems}, 2014, pp. 1988--1996.

\bibitem{ding2015robust}
C.~Ding and D.~Tao, ``Robust face recognition via multimodal deep face
  representation,'' \emph{IEEE Transactions on Multimedia}, vol.~17, no.~11,
  pp. 2049--2058, 2015.

\bibitem{rasiwasia2010new}
N.~Rasiwasia, J.~Costa~Pereira, E.~Coviello, G.~Doyle, G.~R. Lanckriet,
  R.~Levy, and N.~Vasconcelos, ``A new approach to cross-modal multimedia
  retrieval,'' in \emph{Proceedings of the 18th ACM international conference on
  Multimedia}.\hskip 1em plus 0.5em minus 0.4em\relax ACM, 2010, pp. 251--260.

\bibitem{kang2015learning}
C.~Kang, S.~Xiang, S.~Liao, C.~Xu, and C.~Pan, ``Learning consistent feature
  representation for cross-modal multimedia retrieval,'' \emph{IEEE
  Transactions on Multimedia}, vol.~17, no.~3, pp. 370--381, 2015.

\bibitem{greenspan2016guest}
H.~Greenspan, B.~Van~Ginneken, and R.~M. Summers, ``Guest editorial deep
  learning in medical imaging: Overview and future promise of an exciting new
  technique,'' \emph{IEEE Transactions on Medical Imaging}, vol.~35, no.~5, pp.
  1153--1159, 2016.

\bibitem{ding2015deep}
X.~Ding, Y.~Zhang, T.~Liu, and J.~Duan, ``Deep learning for event-driven stock
  prediction,'' in \emph{Twenty-fourth international joint conference on
  artificial intelligence}, 2015.

\bibitem{scholkopf2001learning}
B.~Scholkopf and A.~J. Smola, \emph{Learning with kernels: support vector
  machines, regularization, optimization, and beyond}.\hskip 1em plus 0.5em
  minus 0.4em\relax MIT press, 2001.

\bibitem{guo2016improved}
J.~Guo, ``An improved incremental training approach for large scaled dataset
  based on support vector machine,'' in \emph{2016 IEEE/ACM 3rd International
  Conference on Big Data Computing Applications and Technologies
  (BDCAT)}.\hskip 1em plus 0.5em minus 0.4em\relax IEEE, 2016, pp. 149--157.

\bibitem{liaw2002classification}
A.~Liaw, M.~Wiener \emph{et~al.}, ``Classification and regression by
  randomforest,'' \emph{R news}, vol.~2, no.~3, pp. 18--22, 2002.

\bibitem{deng2009imagenet}
J.~Deng, W.~Dong, R.~Socher, L.-J. Li, K.~Li, and L.~Fei-Fei, ``Imagenet: A
  large-scale hierarchical image database,'' in \emph{2009 IEEE conference on
  computer vision and pattern recognition}.\hskip 1em plus 0.5em minus
  0.4em\relax Ieee, 2009, pp. 248--255.

\bibitem{he2015delving}
K.~He, X.~Zhang, S.~Ren, and J.~Sun, ``Delving deep into rectifiers: Surpassing
  human-level performance on imagenet classification,'' in \emph{Proceedings of
  the IEEE international conference on computer vision}, 2015, pp. 1026--1034.

\bibitem{torrey2010transfer}
L.~Torrey and J.~Shavlik, ``Transfer learning,'' in \emph{Handbook of research
  on machine learning applications and trends: algorithms, methods, and
  techniques}.\hskip 1em plus 0.5em minus 0.4em\relax IGI Global, 2010, pp.
  242--264.

\bibitem{biederman1987recognition}
I.~Biederman, ``Recognition-by-components: a theory of human image
  understanding.'' \emph{Psychological review}, vol.~94, no.~2, p. 115, 1987.

\bibitem{lampert2009learning}
C.~H. Lampert, H.~Nickisch, and S.~Harmeling, ``Learning to detect unseen
  object classes by between-class attribute transfer,'' in \emph{2009 IEEE
  Conference on Computer Vision and Pattern Recognition}.\hskip 1em plus 0.5em
  minus 0.4em\relax IEEE, 2009, pp. 951--958.

\bibitem{frome2013devise}
A.~Frome, G.~S. Corrado, J.~Shlens, S.~Bengio, J.~Dean, T.~Mikolov
  \emph{et~al.}, ``Devise: A deep visual-semantic embedding model,'' in
  \emph{Advances in neural information processing systems}, 2013, pp.
  2121--2129.

\bibitem{lampert2014attribute}
C.~H. Lampert, H.~Nickisch, and S.~Harmeling, ``Attribute-based classification
  for zero-shot visual object categorization,'' \emph{IEEE Transactions on
  Pattern Analysis and Machine Intelligence}, vol.~36, no.~3, pp. 453--465,
  2014.

\bibitem{shigeto2015ridge}
Y.~Shigeto, I.~Suzuki, K.~Hara, M.~Shimbo, and Y.~Matsumoto, ``Ridge
  regression, hubness, and zero-shot learning,'' in \emph{Joint European
  Conference on Machine Learning and Knowledge Discovery in Databases}.\hskip
  1em plus 0.5em minus 0.4em\relax Springer, 2015, pp. 135--151.

\bibitem{wang2015zero}
Z.~Wang, R.~Hu, C.~Liang, Y.~Yu, J.~Jiang, M.~Ye, J.~Chen, and Q.~Leng,
  ``Zero-shot person re-identification via cross-view consistency,'' \emph{IEEE
  Transactions on Multimedia}, vol.~18, no.~2, pp. 260--272, 2015.

\bibitem{bucher2016improving}
M.~Bucher, S.~Herbin, and F.~Jurie, ``Improving semantic embedding consistency
  by metric learning for zero-shot classiffication,'' in \emph{European
  Conference on Computer Vision}.\hskip 1em plus 0.5em minus 0.4em\relax
  Springer, 2016, pp. 730--746.

\bibitem{zhang2016zero}
Z.~Zhang and V.~Saligrama, ``Zero-shot learning via joint latent similarity
  embedding,'' in \emph{Proceedings of the IEEE Conference on Computer Vision
  and Pattern Recognition}, 2016, pp. 6034--6042.

\bibitem{changpinyo2016synthesized}
S.~Changpinyo, W.-L. Chao, B.~Gong, and F.~Sha, ``Synthesized classifiers for
  zero-shot learning,'' in \emph{Proceedings of the IEEE Conference on Computer
  Vision and Pattern Recognition}, 2016, pp. 5327--5336.

\bibitem{kodirov2017semantic}
E.~Kodirov, T.~Xiang, and S.~Gong, ``Semantic autoencoder for zero-shot
  learning,'' in \emph{Proceedings of the IEEE Conference on Computer Vision
  and Pattern Recognition}, 2017, pp. 3174--3183.

\bibitem{zhu2018generative}
Y.~Zhu, M.~Elhoseiny, B.~Liu, X.~Peng, and A.~Elgammal, ``A generative
  adversarial approach for zero-shot learning from noisy texts,'' in
  \emph{Proceedings of the IEEE Conference on Computer Vision and Pattern
  Recognition (CVPR)}, 2018.

\bibitem{guo2019ams}
J.~Guo and S.~Guo, ``Ams-sfe: Towards an alignment of manifold structures via
  semantic feature expansion for zero-shot learning,'' in \emph{2019 IEEE
  International Conference on Multimedia and Expo (ICME)}, 2019, pp. 73--78.

\bibitem{fu2015transductive}
Y.~Fu, T.~M. Hospedales, T.~Xiang, and S.~Gong, ``Transductive multi-view
  zero-shot learning,'' \emph{IEEE transactions on pattern analysis and machine
  intelligence}, vol.~37, no.~11, pp. 2332--2345, 2015.

\bibitem{palatucci2009zero}
M.~Palatucci, D.~Pomerleau, G.~E. Hinton, and T.~M. Mitchell, ``Zero-shot
  learning with semantic output codes,'' in \emph{Advances in neural
  information processing systems}, 2009, pp. 1410--1418.

\bibitem{akata2015evaluation}
Z.~Akata, S.~Reed, D.~Walter, H.~Lee, and B.~Schiele, ``Evaluation of output
  embeddings for fine-grained image classification,'' in \emph{Computer Vision
  and Pattern Recognition (CVPR), 2015 IEEE Conference on}.\hskip 1em plus
  0.5em minus 0.4em\relax IEEE, 2015, pp. 2927--2936.

\bibitem{romera2015embarrassingly}
B.~Romera-Paredes and P.~Torr, ``An embarrassingly simple approach to zero-shot
  learning,'' in \emph{International Conference on Machine Learning}, 2015, pp.
  2152--2161.

\bibitem{akata2016label}
Z.~Akata, F.~Perronnin, Z.~Harchaoui, and C.~Schmid, ``Label-embedding for
  image classification,'' \emph{IEEE transactions on pattern analysis and
  machine intelligence}, vol.~38, no.~7, pp. 1425--1438, 2016.

\bibitem{xian2016latent}
Y.~Xian, Z.~Akata, G.~Sharma, Q.~Nguyen, M.~Hein, and B.~Schiele, ``Latent
  embeddings for zero-shot classification,'' in \emph{Proceedings of the IEEE
  Conference on Computer Vision and Pattern Recognition}, 2016, pp. 69--77.

\bibitem{socher2013zero}
R.~Socher, M.~Ganjoo, C.~D. Manning, and A.~Ng, ``Zero-shot learning through
  cross-modal transfer,'' in \emph{Advances in neural information processing
  systems}, 2013, pp. 935--943.

\bibitem{mikolov2013distributed}
T.~Mikolov, I.~Sutskever, K.~Chen, G.~S. Corrado, and J.~Dean, ``Distributed
  representations of words and phrases and their compositionality,'' in
  \emph{Advances in neural information processing systems}, 2013, pp.
  3111--3119.

\bibitem{ba2015predicting}
L.~J. Ba, K.~Swersky, S.~Fidler, and R.~Salakhutdinov, ``Predicting deep
  zero-shot convolutional neural networks using textual descriptions.'' in
  \emph{ICCV}, 2015, pp. 4247--4255.

\bibitem{zhang2017learning}
L.~Zhang, T.~Xiang, S.~Gong \emph{et~al.}, ``Learning a deep embedding model
  for zero-shot learning,'' 2017.

\bibitem{changpinyo2016predicting}
S.~Changpinyo, W.-L. Chao, and F.~Sha, ``Predicting visual exemplars of unseen
  classes for zero-shot learning,'' \emph{arXiv preprint arXiv:1605.08151},
  2016.

\bibitem{lu2015unsupervised}
Y.~Lu, ``Unsupervised learning on neural network outputs: with application in
  zero-shot learning,'' \emph{arXiv preprint arXiv:1506.00990}, 2015.

\bibitem{zhang2015zero}
Z.~Zhang and V.~Saligrama, ``Zero-shot learning via semantic similarity
  embedding,'' in \emph{Proceedings of the IEEE International Conference on
  Computer Vision}, 2015, pp. 4166--4174.

\bibitem{fu2015zero}
Z.~Fu, T.~Xiang, E.~Kodirov, and S.~Gong, ``Zero-shot object recognition by
  semantic manifold distance,'' in \emph{Proceedings of the IEEE Conference on
  Computer Vision and Pattern Recognition}, 2015, pp. 2635--2644.

\bibitem{akata2016multi}
Z.~Akata, M.~Malinowski, M.~Fritz, and B.~Schiele, ``Multi-cue zero-shot
  learning with strong supervision,'' in \emph{Proceedings of the IEEE
  Conference on Computer Vision and Pattern Recognition}, 2016, pp. 59--68.

\bibitem{long2017zero}
Y.~Long, L.~Liu, L.~Shao, F.~Shen, G.~Ding, and J.~Han, ``From zero-shot
  learning to conventional supervised classification: Unseen visual data
  synthesis,'' 2017.

\bibitem{li2017zero}
Y.~Li, D.~Wang, H.~Hu, Y.~Lin, and Y.~Zhuang, ``Zero-shot recognition using
  dual visual-semantic mapping paths,'' in \emph{IEEE Conference on Computer
  Vision and Pattern Recognition (CVPR)}, 2017, pp. 5207--5215.

\bibitem{song2018transductive}
J.~Song, C.~Shen, Y.~Yang, Y.~Liu, and M.~Song, ``Transductive unbiased
  embedding for zero-shot learning,'' in \emph{Proceedings of the IEEE
  Conference on Computer Vision and Pattern Recognition}, 2018, pp. 1024--1033.

\bibitem{AAAI1816805}
\BIBentryALTinterwordspacing
B.~Tong, M.~Klinkigt, J.~Chen, X.~Cui, Q.~Kong, T.~Murakami, and Y.~Kobayashi,
  ``Adversarial zero-shot learning with semantic augmentation,'' 2018.
  [Online]. Available:
  \url{https://www.aaai.org/ocs/index.php/AAAI/AAAI18/paper/view/16805}
\BIBentrySTDinterwordspacing

\bibitem{wang2017quantifying}
Q.~Wang, M.~Chen, and X.~Li, ``Quantifying and detecting collective motion by
  manifold learning.'' in \emph{AAAI}, 2017, pp. 4292--4298.

\bibitem{xu2017matrix}
X.~Xu, F.~Shen, Y.~Yang, D.~Zhang, H.~T. Shen, and J.~Song, ``Matrix
  tri-factorization with manifold regularizations for zero-shot learning,'' in
  \emph{Proceeding of the IEEE conference on computer vision and pattern
  recognition. CVPR}, 2017.

\bibitem{goodfellow2014generative}
I.~Goodfellow, J.~Pouget-Abadie, M.~Mirza, B.~Xu, D.~Warde-Farley, S.~Ozair,
  A.~Courville, and Y.~Bengio, ``Generative adversarial nets,'' in
  \emph{Advances in neural information processing systems}, 2014, pp.
  2672--2680.

\bibitem{baldi2012autoencoders}
P.~Baldi, ``Autoencoders, unsupervised learning, and deep architectures,'' in
  \emph{Proceedings of ICML workshop on unsupervised and transfer learning},
  2012, pp. 37--49.

\bibitem{zhang2018stackgan++}
H.~Zhang, T.~Xu, H.~Li, S.~Zhang, X.~Wang, X.~Huang, and D.~N. Metaxas,
  ``Stackgan++: Realistic image synthesis with stacked generative adversarial
  networks,'' \emph{IEEE transactions on pattern analysis and machine
  intelligence}, vol.~41, no.~8, pp. 1947--1962, 2018.

\bibitem{vincent2010stacked}
P.~Vincent, H.~Larochelle, I.~Lajoie, Y.~Bengio, and P.-A. Manzagol, ``Stacked
  denoising autoencoders: Learning useful representations in a deep network
  with a local denoising criterion,'' \emph{Journal of machine learning
  research}, vol.~11, no. Dec, pp. 3371--3408, 2010.

\bibitem{yu2013embedding}
W.~Yu, G.~Zeng, P.~Luo, F.~Zhuang, Q.~He, and Z.~Shi, ``Embedding with
  autoencoder regularization,'' in \emph{Joint European Conference on Machine
  Learning and Knowledge Discovery in Databases}.\hskip 1em plus 0.5em minus
  0.4em\relax Springer, 2013, pp. 208--223.

\bibitem{chu2017stacked}
W.~Chu and D.~Cai, ``Stacked similarity-aware autoencoders,'' in
  \emph{Proceedings of the 26th International Joint Conference on Artificial
  Intelligence}.\hskip 1em plus 0.5em minus 0.4em\relax AAAI Press, 2017, pp.
  1561--1567.

\bibitem{guo2019ee}
J.~Guo and S.~Guo, ``Ee-ae: An exclusivity enhanced unsupervised feature
  learning approach,'' in \emph{ICASSP 2019-2019 IEEE International Conference
  on Acoustics, Speech and Signal Processing (ICASSP)}.\hskip 1em plus 0.5em
  minus 0.4em\relax IEEE, 2019, pp. 3517--3521.

\bibitem{DBLP:journals/corr/KingmaW13}
\BIBentryALTinterwordspacing
D.~P. Kingma and M.~Welling, ``Auto-encoding variational bayes,'' in \emph{2nd
  International Conference on Learning Representations, {ICLR} 2014, Banff, AB,
  Canada, April 14-16, 2014, Conference Track Proceedings}, 2014. [Online].
  Available: \url{http://arxiv.org/abs/1312.6114}
\BIBentrySTDinterwordspacing

\bibitem{blei2017variational}
D.~M. Blei, A.~Kucukelbir, and J.~D. McAuliffe, ``Variational inference: A
  review for statisticians,'' \emph{Journal of the American statistical
  Association}, vol. 112, no. 518, pp. 859--877, 2017.

\bibitem{chonavel2003fast}
T.~Chonavel, B.~Champagne, and C.~Riou, ``Fast adaptive eigenvalue
  decomposition: a maximum likelihood approach,'' \emph{Signal processing},
  vol.~83, no.~2, pp. 307--324, 2003.

\bibitem{wah2011caltech}
C.~Wah, S.~Branson, P.~Welinder, P.~Perona, and S.~Belongie, ``The caltech-ucsd
  birds-200-2011 dataset,'' 2011.

\bibitem{farhadi2009describing}
A.~Farhadi, I.~Endres, D.~Hoiem, and D.~Forsyth, ``Describing objects by their
  attributes,'' in \emph{Computer Vision and Pattern Recognition, 2009. CVPR
  2009. IEEE Conference on}.\hskip 1em plus 0.5em minus 0.4em\relax IEEE, 2009,
  pp. 1778--1785.

\bibitem{patterson2014sun}
G.~Patterson, C.~Xu, H.~Su, and J.~Hays, ``The sun attribute database: Beyond
  categories for deeper scene understanding,'' \emph{International Journal of
  Computer Vision}, vol. 108, no. 1-2, pp. 59--81, 2014.

\bibitem{russakovsky2015imagenet}
O.~Russakovsky, J.~Deng, H.~Su, J.~Krause, S.~Satheesh, S.~Ma, Z.~Huang,
  A.~Karpathy, A.~Khosla, M.~Bernstein \emph{et~al.}, ``Imagenet large scale
  visual recognition challenge,'' \emph{International Journal of Computer
  Vision}, vol. 115, no.~3, pp. 211--252, 2015.

\bibitem{everingham2008pascal}
M.~Everingham, L.~Van~Gool, C.~K. Williams, J.~Winn, and A.~Zisserman, ``The
  pascal visual object classes challenge 2007 (voc 2007) results (2007),''
  2008.

\bibitem{szegedy2015going}
C.~Szegedy, W.~Liu, Y.~Jia, P.~Sermanet, S.~Reed, D.~Anguelov, D.~Erhan,
  V.~Vanhoucke, A.~Rabinovich \emph{et~al.}, ``Going deeper with
  convolutions.''\hskip 1em plus 0.5em minus 0.4em\relax Cvpr, 2015.

\bibitem{hsu2003practical}
C.-W. Hsu, C.-C. Chang, C.-J. Lin \emph{et~al.}, ``A practical guide to support
  vector classification.''

\bibitem{kendall2018multi}
A.~Kendall, Y.~Gal, and R.~Cipolla, ``Multi-task learning using uncertainty to
  weigh losses for scene geometry and semantics,'' in \emph{Proceedings of the
  IEEE Conference on Computer Vision and Pattern Recognition}, 2018, pp.
  7482--7491.

\bibitem{sener2018multi}
O.~Sener and V.~Koltun, ``Multi-task learning as multi-objective
  optimization,'' in \emph{Advances in Neural Information Processing Systems},
  2018, pp. 527--538.

\bibitem{guo2019adaptive}
J.~Guo and S.~Guo, ``Adaptive adjustment with semantic feature space for
  zero-shot recognition,'' in \emph{ICASSP 2019-2019 IEEE International
  Conference on Acoustics, Speech and Signal Processing (ICASSP)}.\hskip 1em
  plus 0.5em minus 0.4em\relax IEEE, 2019, pp. 3287--3291.

\bibitem{yang2014unified}
Y.~Yang and T.~M. Hospedales, ``A unified perspective on multi-domain and
  multi-task learning,'' \emph{arXiv preprint arXiv:1412.7489}, 2014.

\bibitem{fu2016semi}
Y.~Fu and L.~Sigal, ``Semi-supervised vocabulary-informed learning,'' in
  \emph{Proceedings of the IEEE Conference on Computer Vision and Pattern
  Recognition}, 2016, pp. 5337--5346.

\bibitem{yang2018learning}
F.~S.~Y. Yang, L.~Zhang, T.~Xiang, P.~H. Torr, and T.~M. Hospedales, ``Learning
  to compare: Relation network for few-shot learning.''

\bibitem{rahman2018unified}
S.~Rahman, S.~Khan, and F.~Porikli, ``A unified approach for conventional
  zero-shot, generalized zero-shot and few-shot learning,'' \emph{IEEE
  Transactions on Image Processing}, 2018.

\bibitem{yu2018zero}
Y.~Yu, Z.~Ji, J.~Guo, and Z.~Zhang, ``Zero-shot learning via latent space
  encoding,'' \emph{IEEE transactions on cybernetics}, no.~99, pp. 1--12, 2018.

\bibitem{maaten2008visualizing}
L.~v.~d. Maaten and G.~Hinton, ``Visualizing data using t-sne,'' \emph{Journal
  of machine learning research}, vol.~9, no. Nov, pp. 2579--2605, 2008.

\end{thebibliography}

\end{document}